\documentclass[journal]{IEEEtran}
\usepackage{amsmath,amssymb,amsfonts}
\usepackage{algorithmic}
\usepackage{algorithm}
\usepackage{array}
\usepackage{textcomp}
\usepackage{stfloats}
\usepackage{url}
\usepackage{verbatim}
\usepackage{graphicx}
\usepackage{cite}
\usepackage{booktabs}  % for lines
\usepackage{multirow}  % for multirow tables
\usepackage{soul}  % for underlining
\usepackage{tikz}  % for embedded graphs
\usepackage{pgfplots}  % for embedded graphs
\usepackage{bm}  % for bold greek symbols
\usepackage[labelformat=simple]{subcaption}
\DeclareCaptionLabelSeparator{periodspace}{.\quad}
\pgfplotsset{compat=1.18}
\usepackage{enumitem}  % for fancy list
\usepackage{hyperref}  % for links
\usepackage{orcidlink} % for orcid links
\usepackage{balance}

\DeclareMathOperator*{\argmax}{arg\,max}
\begin{document}
\bstctlcite{BSTcontrol}

\title{T-TAME: Trainable Attention Mechanism for Explaining Convolutional Networks and Vision Transformers}

\author{
\IEEEauthorblockN{
Mariano V. Ntrougkas \orcidlink{0009-0004-0569-0837}\IEEEauthorrefmark{1},
Nikolaos Gkalelis\orcidlink{0000-0001-6741-3334}\IEEEauthorrefmark{1}, 
Vasileios Mezaris\orcidlink{0000-0002-0121-4364}\IEEEauthorrefmark{1}}\\
\IEEEauthorblockA{\IEEEauthorrefmark{1}Centre for Research and Technology Hellas (CERTH) / Information Technologies Institute (ITI) \\ Thermi 57001, Greece. \\ \{ntrougkas,gkalelis,bmezaris\}@iti.gr}
\thanks{This work was supported by the EU Horizon 2020 programme under grant agreement H2020-101021866 CRiTERIA.}}

\IEEEoverridecommandlockouts

\maketitle

\begin{abstract}
The development and adoption of Vision Transformers and other deep-learning architectures for image classification tasks has been rapid. However, the ``black box'' nature of neural networks is a barrier to adoption in applications where explainability is essential. While some techniques for generating explanations have been proposed, primarily for Convolutional Neural Networks, adapting such techniques to the new paradigm of Vision Transformers is non-trivial. 
This paper presents T-TAME, \emph{Transformer-compatible Trainable Attention Mechanism for Explanations}\footnote{Source code and trained explainability models will be made publicly available upon publication.}, a general methodology for explaining deep neural networks used in image classification tasks.
The proposed architecture and training technique can be easily applied to any convolutional or Vision Transformer-like neural network, using a streamlined training approach.
After training, explanation maps can be computed in a single forward pass; these explanation maps are comparable to or outperform the outputs of computationally expensive perturbation-based explainability techniques, achieving SOTA performance.
We apply T-TAME to three popular deep learning classifier architectures, VGG-16, ResNet-50, and ViT-B-16, trained on the ImageNet dataset, and we demonstrate improvements over existing state-of-the-art explainability methods. A detailed analysis of the results and an ablation study
provide insights into how the T-TAME design choices affect the quality of the generated explanation maps. 
\end{abstract}


\section{Introduction}
\label{sec:introduction}

\begin{figure}[tbp]%
\centering
\includegraphics[width=3.3in]{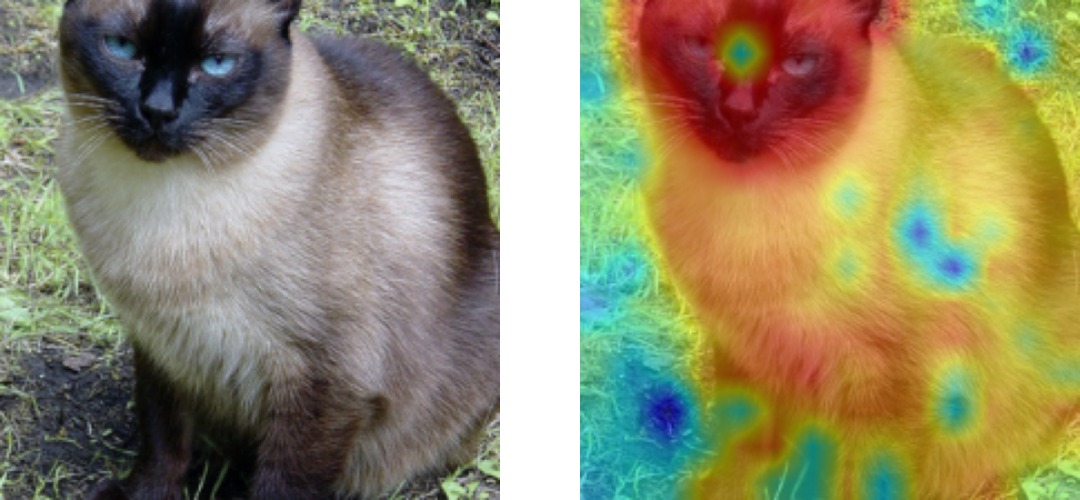}
\caption{An explanation produced by T-TAME for the ViT-B-16 backbone classifier. The input image (left) belongs to the class ``Siamese cat'' and is correctly classifier by ViT-B-16. The produced explanation (right) highlights the salient features of the image that explain the decision of this specific classifier (areas in red color in the explanation map), which do not necessarily coincide with the image region where the human-recognizable ``Siamese cat'' object appears. In this example, the explanation map reveals that it is primarily the cat's head that this classifier relied on to render its decision.
}
\label{fig:XaiVsLoc}
\end{figure}%

\IEEEPARstart{V}{ision} Transformers (ViTs) \cite{vit} have been found to match or outperform Convolutional Neural Networks (CNNs) in many important visual tasks such as natural image classification \cite{touvron2021training}, classification of masses in breast ultrasound \cite{breast}, skin cancer classification \cite{skin}, and face recognition \cite{face}. As a result of the complex multi-layer nonlinear structure and end-to-end learning strategy of models, such as CNNs and ViTs, they typically act as ``black box'' models that lack transparency \cite{HamonIEEECLM2022}. This fact makes it difficult to convince users in critical fields, such as healthcare, law, and governance to trust and employ such systems \cite{gorski2021explainable}, thus limiting the adoption of Artificial Intelligence \cite{amann2020explainability,HamonIEEECLM2022}. Therefore, it is necessary to develop solutions that address the transparency challenge of deep neural networks.
 
Explainable artificial intelligence (XAI) is an active research area in the field of machine learning. XAI focuses on developing explainable techniques that help users of AI systems comprehend, trust, and more efficiently manage them \cite{arrieta2020explainable,samek2021explaining}. 
\IEEEpubidadjcol
For the image classification task, several explanation approaches have been proposed to tackle the explainability problem for CNN and ViT models \cite{samek2021explaining}.
These methods typically produce an explanation map, also referred to as a saliency map, highlighting the salient input features. We must stress that explainability methods should not be confused with approaches targeting weakly supervised learning tasks such as weakly supervised object localization or segmentation  \cite{JiangTPAMI2021}, which also generate superficially similar heatmaps as an intermediate step. Contrary to the latter, the goal of explainability approaches is to explain the classifier's decision rather than to locate the region of the target object (for an example, see Fig. \ref{fig:XaiVsLoc}).

The existing explanation approaches for image classifiers can be roughly categorized as follows. Gradient-based methods, such as Grad-CAM and Grad-CAM++, were pioneering approaches in explaining CNNs \cite{selvaraju2017grad,chattopadhay2018grad} and were also among the first methods applied to ViTs \cite{Chefer_2021_CVPR}.
Since these approaches utilize gradient information, they are subject to associated limitations such as gradient saturation and noise issues, resulting in explanations that may include high-frequency variations \cite{AdebayoNIPS2018,Kindermans2019}.
Relevance-based methods, on the other hand, use a Taylor decomposition of a relevance function they define to propagate relevance of pixel information through the examined network \cite{MONTAVON2017211, bach2015pixel,Chefer_2021_CVPR}. 
These methods do not directly rely on gradient information and are therefore less prone to the limitations associated with gradient-based approaches; however, their difficulty to be adapted to novel classifier architectures restricts their applicability \cite{Holzinger2022}.
Finally, perturbation- \cite{petsiuk2018rise,wang2020score} and response-based approaches \cite{sattarzadeh2021explaining,sudhakar2021ada,englebert2022backward,barkan2023visual} observe the output's sensitivity to a multitude of small input changes, and combine intermediate network representations to derive an explanation, respectively.
The methods within these categories operate without using gradients and thus avoid relevant drawbacks; however, their process for generating an explanation is computationally very expensive.

Distinguished from the above works, L-CAM \cite{Gkartzonika2022} is a trainable response-based method: it utilizes an appropriate objective function to guide the training of an attention mechanism in order to derive explanation maps of high quality in one forward pass.
However, L-CAM uses the feature maps of only the last convolutional layer of the frozen CNN model to be explained (hereafter also referred to as the ``backbone network''), thus may not be able to adequately capture the information used within this backbone for making a classification decision. 
Additionally, L-CAM is not applicable to ViTs, because ViT feature maps are not three-dimensional, unlike CNN feature maps, and because of the different ways in which ViTs handle input perturbations (see \cite{vitrobust} for a comparison w.r.t. robustness between ViTs and CNNs).

To this end, we propose T-TAME: Transformer-compatible Trainable Attention Mechanism for Explanations. T-TAME is inspired by the learning-based paradigm of L-CAM. Unlike L-CAM, T-TAME exploits intermediate feature maps extracted from multiple layers of the backbone network. These features are then used to train a multi-branch hierarchical attention architecture for generating class-specific explanation maps in a single forward pass. Additionally, T-TAME introduces components that manage the compatibility of the trainable attention mechanism with the backbone network, enabling its use with both CNN and ViT backbones. We demonstrate that T-TAME generates higher quality explanation maps over the current SOTA explainability methods, by performing a rich set of qualitative and quantitative comparisons. A preliminary version of this work, still applicable only to CNN backbones, was presented in \cite{TAME}.

In summary, the contributions of this paper are:
\begin{itemize}
    \item We present the first, to the best of our knowledge, trainable post-hoc method for generating explanation maps for both CNN and Transformer-based image classification networks, which utilizes an attention mechanism to process feature maps from multiple layers.
    \item We provide a comprehensive evaluation study of the proposed T-TAME method for three heterogeneous backbones: the widely used CNN models VGG-16 \cite{simonyan2014very} and ResNet-50 \cite{he2016deep}, as well as the breakthrough ViT model ViT-B-16 \cite{vit}.
    \item Based on example explanations produced by T-TAME and ablation experiments, we gain insights into the ViT classifier. Specifically, we demonstrate ViT's global view of input images, thanks to its multi-head attention layer, and we confirm its robustness to out-of-sample distributions of input images.
    
\end{itemize}

\section{Related Work}
\label{s:RelatedWork}

We start by briefly discussing the broader domain of XAI. 
The ability to provide an explanation for why a specific decision was made is now seen as a desirable feature of intelligent systems \cite{confalonieri2021historical}. These explanations serve to help users understand the AI system's underlying model, facilitating its effective use and maintenance. Additionally, they assist users in identifying and correcting errors in the AI system's outputs, thus aiding in debugging. Furthermore, explanations can be used for educational purposes, helping users to explore and understand new concepts within a particular domain. Finally, explanations contribute to users' trust and cogency by offering actionable insights and convincing them that the system's decisions can be trusted.

What constitutes a ``good'' explanation for an AI system is still an open research question. Three important properties for explanations have been identified by social science research on how humans explain their decisions to each other \cite{miller2019explanation}; here we briefly discuss how the current paradigm of explanation methods for vision classifiers aligns with these properties. First, explanations are counterfactual; they justify a decision in opposition to other choices, i.e., why a backbone network classified a specific image as a certain class instead of another possible class. An explanation method can be counterfactual by providing explanation maps for each class that is considered by the backbone network, thus, allowing the user to compare explanation maps for different possible classification decisions. Second, explanations are selected in a biased manner, so as to not overwhelm the user with information. To this end, in the vision classifier domain, the most common form of explanation 
is
a heatmap (a.k.a. explanation map). Third, explanations are social, thus they need to align with the mental model of the user of an AI system. When a user views an image, typically they pay more attention to some parts of the image than to others. In a direct analogy, the user would expect an image classification model to focus more or less on specific regions of the input image for making its classification decision; these are the image regions that are highlighted by the explanation map.

There is a wide range of explainability methods, which are often also referred to as feature attribution methods.
Based on the scope of their explanations, i.e., whether they are used to produce explanations for single predictions or for the overall model, these methods can be characterized as local or global \cite{globalexp}.
Another important distinction regarding an explainability method arises from its relationship with the model it aims to explain, classifying it as either ante-hoc or post-hoc.
The former approaches require architectural modifications that have to be applied prior to the training of the classifier. Several intrinsically explainable classifiers that fall in this category have been developed \cite{bohle2023holistically}. Contrarily, a method that can be directly applied to an already-trained classifier is a post-hoc method.
Post-hoc explainability approaches can be applied to existing off-the-shelf classifiers, thus providing users with the freedom to choose a top-performing classifier without compromising on model explainability \cite{CrookREW23}.
These approaches can be further categorized as model-specific or model-agnostic, depending on whether they are applicable to only specific models or any type of model.
For a more comprehensive review of the different taxonomies of explanation methods and the different approaches therein, the interested reader is referred to \cite{arrieta2020explainable,samek2021explaining, ChamolaIEEEAccess23, hassija2024interpreting}.

Among the above-described classes of explainability methods, local post-hoc methods are most widely applicable to the task of explaining deep learning-based image classification models.
In the following, we survey the state-of-the-art approaches in this category that are most closely related to ours.
These approaches can be roughly categorized into gradient-, relevance-, perturbation- and response-based.
Gradient-based methods \cite{selvaraju2017grad,chattopadhay2018grad} compute the gradient of a given input with backpropagation and modify it in various ways to produce an explanation map.
Grad-CAM \cite{selvaraju2017grad}, one of the first in this category, uses global average pooling in the gradients of the backbone network's logits with respect to the feature maps to compute weights.
The explanation maps are obtained as the weighted combination of feature maps, using the computed weights. Grad-CAM++ \cite{chattopadhay2018grad} similarly uses gradients to generate explanation maps.
These methods suffer from the same issues as the gradients they use: neural network gradients can be noisy and suffer from saturation problems for typical activation functions such as ReLU, GELU, and Sigmoid \cite{AdebayoNIPS2018}.

Relevance-based methods \cite{MONTAVON2017211, bach2015pixel, Chefer_2021_CVPR} use a Taylor approximation of the gradients to propagate relevance of pixel information through the examined network. The propagation function is a modified version of backpropagation, aimed at reducing noise and retaining layer-wise salient information. Relevance is propagated to the input image, producing an explanation map. An early work of this class,
Deep Taylor Decomposition (DTD) \cite{MONTAVON2017211}, directly uses gradients, propagating them throughout the network and accumulating the contribution to the output prediction from each layer of the network. Layer-wise Relevance Propagation (LRP) \cite{bach2015pixel} cemented the use of Taylor approximation to explain general network architectures. In contrast to methods like Grad-CAM, this method combines information from all of the layers in the network. An extension of the LRP method for Transformer-based architectures, including ViTs, is presented in \cite{Chefer_2021_CVPR}.
However, applying these methods to novel architectures and new network layers is not a straightforward task, requiring the careful fulfillment of the relevance propagation rules through each network operation and dealing with practical issues that may arise, such as numerical instability; thus, their applicability is limited.

Perturbation-based methods \cite{petsiuk2018rise,wang2020score} attempt to repeatedly alter the input and produce explanations based on the observed changes in the confidence of the original prediction; thus, avoid gradient-related problems such as vanishing or noisy gradients.
For instance, RISE \cite{petsiuk2018rise} utilizes Monte Carlo sampling to generate random masks, which are then used to perturb the input image and generate a respective CNN classification score.
Using the computed scores as weights, the explanation map is derived as the weighted combination of the generated random masks.
Score-CAM \cite{wang2020score}, on the other hand, utilizes the feature maps from the final layer of the network as masks by upsampling them to the size of the input image, instead of generating random masks.
Thus, RISE and Score-CAM, as most methods in this category, require many forward passes through the network (in the order of hundreds or thousands) to generate an explanation, considerably increasing the inference time and computational cost.

Response-based methods \cite{sattarzadeh2021explaining,sudhakar2021ada,englebert2022backward,Gkartzonika2022, barkan2023visual} use feature maps or activations of the backbone's layers in the inference stage to interpret the decision-making process of the backbone neural network.
One of the first methods in this category, CAM \cite{zhou2016learning}, uses the output of the backbone's global average pooling layer as weights, and computes the weighted average of the features maps at the final convolutional layer.
CAM requires the presence of such a global average pooling layer in the target model's architecture, restricting its applicability.
SISE \cite{sattarzadeh2021explaining}, and later Ada-SISE \cite{sudhakar2021ada}, aggregate feature maps in a cascading manner to produce explanation maps of any CNN model.
Similarly, Poly-CAM \cite{englebert2022backward} uses feature maps from multiple layers, upscales them to the largest spatial dimension present in the set, and then combines them in a cascading manner. Iterated Integrated Attributions (IIA) \cite{barkan2023visual} is a generalization of Integrated Gradients \cite{method:ig} that further employs gradients from internal feature maps. It is also applied to ViT models by using attention matrices as feature maps; the usage of gradients of the input and feature maps from the last two layers before the classification stage are considered. Similarly to perturbation-based methods, the above methods require either multiple forward passes in the case of SISE, Ada-SISE, and Poly-CAM, or multiple backward passes in the case of IIA, to produce an explanation. 

Finally, the category of trainable response-based explanation methods is represented by L-CAM \cite{Gkartzonika2022}. L-CAM mitigates the limitations of response-based methods by using a learned attention mechanism to compute class-specific explanations in one forward pass.
However, it can only harness the salient information of feature maps from a single layer of a CNN backbone. The proposed T-TAME method is a trainable response-based method that addresses the limitations of L-CAM, by using feature maps from multiple layers and by being applicable to both CNN and Transformer-based architectures.
In contrast to the majority of the approaches described above, which traverse the network multiple times to provide an explanation, the proposed approach is computationally inexpensive at the inference stage, requiring only a single forward pass.

We should also note that the methods of \cite{jetley2018learn, fukui2019attention} take a somewhat similar approach to ours in that they produce explanation maps using an attention mechanism and multiple sets of feature maps.
However, these methods are ante-hoc, jointly training the attention model with the CNN backbone that learns to perform the desired image classification task.
In contrast, T-TAME does not modify the trained target (a.k.a. backbone) model, whose weights remain frozen. I.e., T-TAME is a post-hoc method, exclusively optimizing the attention mechanism in an unsupervised learning manner to generate visual explanations.
Thus, no direct comparisons can be drawn with \cite{jetley2018learn, fukui2019attention} as they provide explanations for a different, concurrently-trained classifier rather than an already optimized backbone. Finally, as T-TAME is based on an attention mechanism, special tribute must be paid to \cite{itti_model_1998} for the first use of hierarchical attention, inspired by early primate vision, in the field of image processing.

\section{Methodology}
\label{sec:tame}
\begin{table}[htbp]
  \centering
  \caption{
  Main symbols used in Section~\ref{sec:tame}.} 
    \begin{tabular}{c}
    Symbols in \textbf{bold} denote tensors or sets. Scalars and operators are \\ denoted in normal font.\\
    \end{tabular}
    \begin{tabular}{ll}
    \toprule
    Symbols & Description\\
    \midrule
    $f()$ & Classifier neural network \\
    $\boldsymbol{I}$ & Input image \\
    $Cls$& Scalar specifying the number of classes on which \\
    &classifier $f$ has been trained on\\
    $C,\,W,\, H$ & Dimensions of input image tensor: number of channels,\\
    &width, height \\
    $N,\, D,\,P$ & Parameters of ViT-based backbone: number of patches, \\ &  hidden size of tokens, width (\& height) of a single patch\\ 
    $\boldsymbol{L}_i$& $i$th feature map \\
    $\boldsymbol{L}^s$ & Set of feature maps extracted from $s$ layers \\
    $\boldsymbol{A}_i$ & Attention map of the $i$th feature branch\\
    $\boldsymbol{A}^s$& Set of attention maps from $s$ feature branches\\
    $C_i,\, W_i,\, H_i$ & Dimensions of the $i$th feature map (same as the $i$th \\
    & attention map): number of channels, width, height \\
    $\boldsymbol{E}$ & Explanation maps (for all $Cls$ classes) \\
    $\boldsymbol{E}_n$ & Explanation map for the $n$th class\\
    $W_e,\, H_e$& Dimensions of explanation map 
    $\boldsymbol{E}_n$: width, height \\
    $\boldsymbol{\Psi}$& Set of explanation maps for a subset of the $Cls$ classes\\
    \bottomrule
    \end{tabular}%
  \label{tab:notation}%
\end{table}%

\begin{figure*}[ht!]%
\centering
\includegraphics[width=0.56\textwidth]{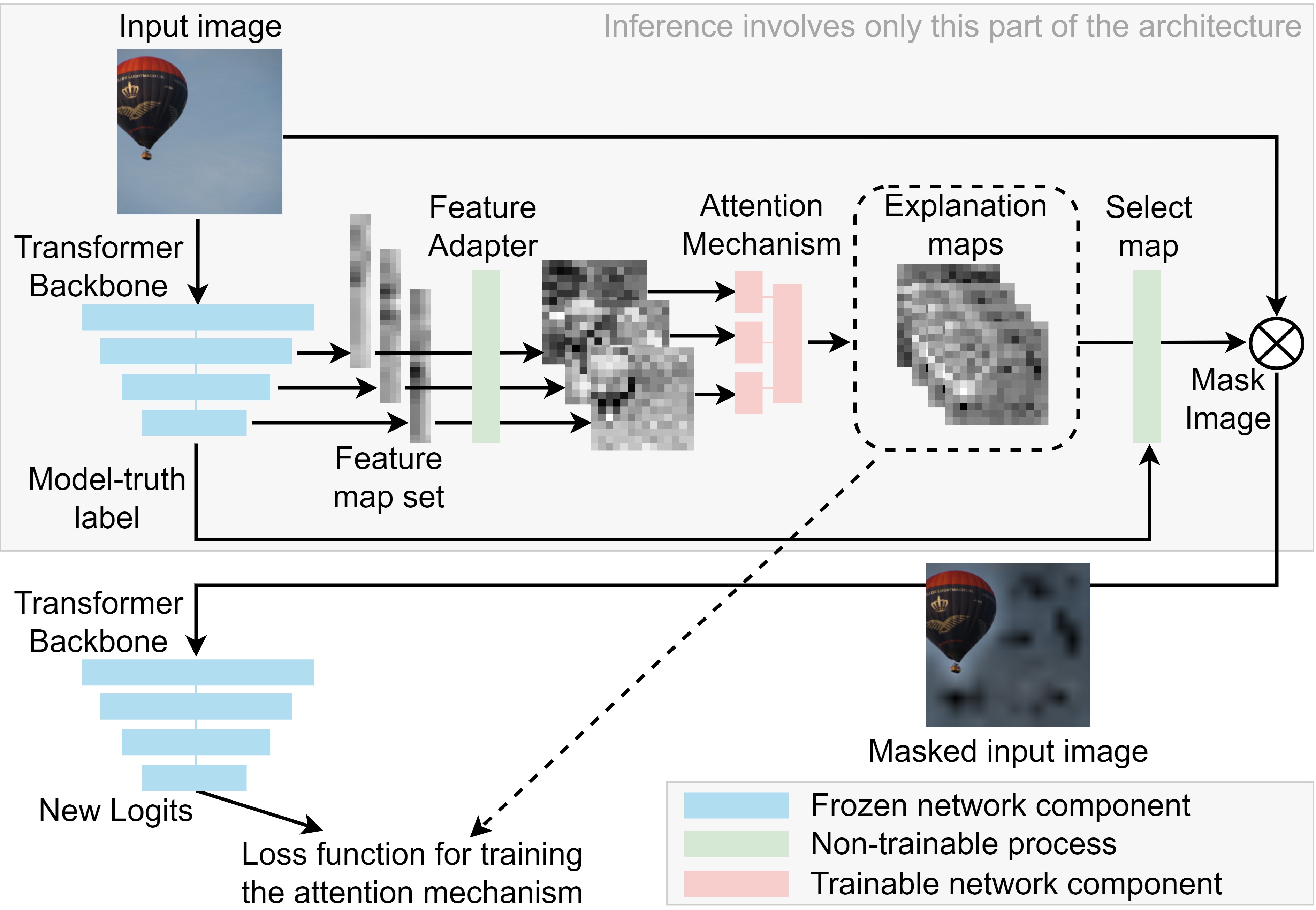}
\caption{Overview of the T-TAME method, showing both the overall architecture used for training the explanation-generating attention mechanism and the inference-stage use of the trained attention mechanism. In this illustration, T-TAME is applied on a ViT backbone.}
\label{fig:overview}
\end{figure*}%

\begin{figure*}[ht!]
\centering
\begin{tabular}{c}
      \includegraphics[width=0.65\textwidth]{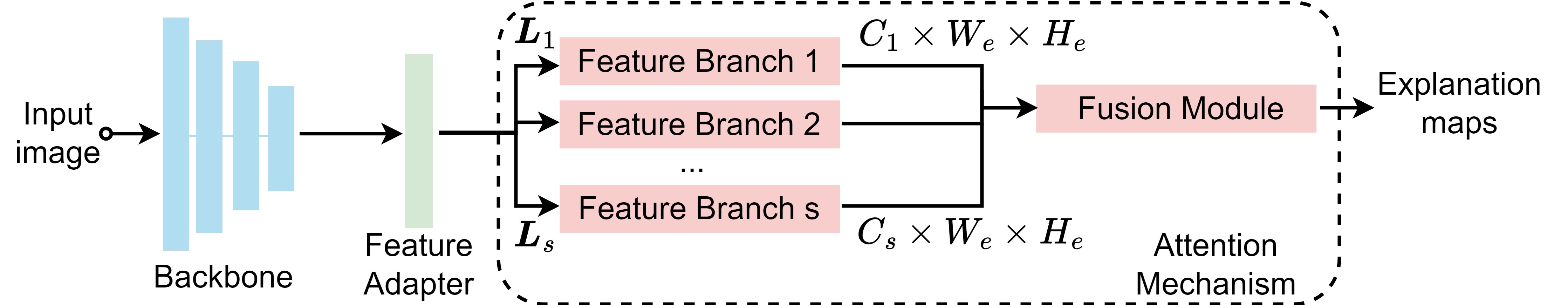}\\
      (a) \\
      \includegraphics[width=0.68 \linewidth]{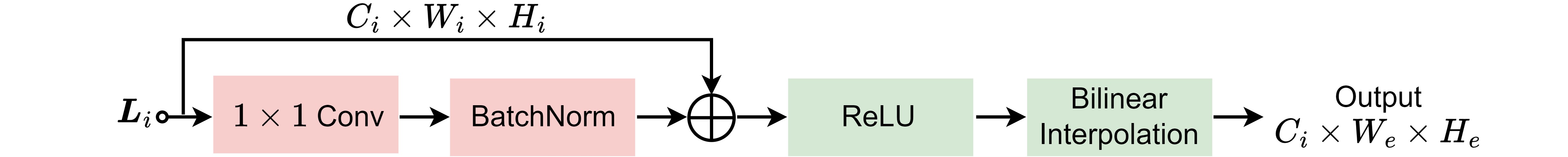} \\
      (b) \\ \\
      \includegraphics[width=0.68\linewidth]{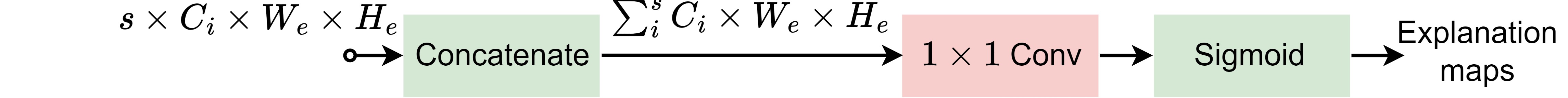} \\
      (c) \\
\end{tabular}
\caption{Structure of the core architecture of the proposed T-TAME method: (a) Overall structure (feature map adapter and attention mechanism), (b) detailed structure of a feature branch of the attention mechanism, (c) detailed structure of the fusion module of the attention mechanism. Color coding retains the same meaning as in Fig.~\ref{fig:overview}.}
\label{fig:general}
\end{figure*}

\subsection{Problem formulation}
\label{ss:problemFormulations}

Let $f$ be a trained backbone network for which we want to generate explanation maps,
\begin{equation}
    f\colon Sp\left(\boldsymbol{I}\right) \to  [0,1]^{Cls},
\end{equation}
where $Sp\left(\boldsymbol{I}\right)$ is the space of three-dimensional input images,
\begin{gather}
    Sp\left(\boldsymbol{I}\right) = \left\{\boldsymbol{I} \mid \boldsymbol{I} \colon \boldsymbol{\Lambda} \to \mathbb{R}\right\}, \\
    \boldsymbol{\Lambda} = \{1, \dotsc, C\} \times \{1, \dotsc, W\} \times \{1, \dotsc, H\}, \nonumber
\end{gather} $C,\, W,\, H \in \mathbb{N}$  are the input image tensor dimensions, i.e., number of channels, width, and height, respectively \cite{sattarzadeh2021explaining,petsiuk2018rise}; and $Cls$ is the number of classes that $f$ has been trained to classify.
E.g., for RGB images in the ImageNet dataset, typically $H=W=224$ is the image height/width, $C=3$ is the number of channels, $Cls=1000$ and the image tensor $\boldsymbol{I}$ is the mapping from the 3D coordinates to pixel values, commonly in the range $\left[0, 1\right]$.

The input image $\boldsymbol{I}$ is transformed to the output $[0,1]^{Cls}$ through various discrete computation steps, called layers. A neural network consists of numerous layers, depending on its specific architecture; a layer's output is referred to as a ``feature map''. Suppose feature maps are extracted from $s$ layers of the backbone network $f$; this set of feature maps is represented as
\begin{equation}
    \boldsymbol{L}^s = \{\boldsymbol{L}_i \mid i \in \{1,\dotsc,s\}\}.
\end{equation}
A feature map $\boldsymbol{L}_i$ of a neural network can take different shapes depending on the type of the backbone network.
For CNNs, a feature map is typically represented as 
\begin{equation}
     \boldsymbol{L}_i \colon \{1, \dotsc, C_i\} \times \{1, \dotsc, W_i\} \times \{1, \dotsc, H_i\} \to \mathbb{R},
     \label{eq:cnn-fm}
\end{equation} where, $C_i, W_i, H_i\in \mathbb{N}$ are the respective channel, width, and height dimensions of the $i$th feature map in the feature map set.
In ViTs \cite{vit}, the feature map 
is represented as
\begin{equation}
  \boldsymbol{L}_i \colon \{1, \dotsc, N + 1\} \times \{1, \dotsc, D\} \to \mathbb{R}, \label{eq:vitFm}  
\end{equation}
where $N$, $D\in \mathbb{N}$ are the number of patches and the constant hidden size through all its layers, respectively.
The former ($N$) equals $H W / P^2$, where $P\in\mathbb{N}$ is the width (\& height) of a single square patch of the input image. $P$, $N$ and $D$ are architecture-dependent values.
For instance, for the ViT-B-16 architecture and input image resolution $W=H=224$, $P=16$, $N=14^2=196$ and $D=768$.
The extra token in the ViT feature map (i.e., the one that increases the map's dimension from $N$ to $N + 1$) is called the ``class token'' and is used by the classification layer. Thus, in CNNs, feature maps are 3D tensors, while in ViTs they are 2D tensors.

Assume an attention mechanism defined as
\begin{equation}
AM \colon Sp\left(\boldsymbol{L}^s\right) \to Sp\left(\boldsymbol{E}\right),    
\end{equation}
where 
\begin{equation}
\boldsymbol{E} \colon  \{1, \dotsc, Cls\} \times\{1, \dotsc, W_e\} \times \{1, \dotsc, H_e\} \to \left[0,1\right] \label{eq:expl-maps}
\end{equation} 
are the explanation maps produced by the attention mechanism, having spatial dimensions $W_e$, $H_e$. $Sp\left(\boldsymbol{L}^s\right)$ and $Sp\left(\boldsymbol{E}\right)$ denote the space of feature map sets and explanation maps, respectively.
The explanations are class-discriminative, i.e., each slice of $\boldsymbol{E}$ along its first dimension, $\boldsymbol{E}_n,\ n\in\{1,\dotsc,Cls\}$, is the explanation map corresponding to the $n$th class on which the classifier $f$ has been trained on.

Given the above general formulation, we propose T-TAME: a trainable attention mechanism architecture, along with a compatible training method. The proposed attention mechanism is applicable to a wide range of classifier backbones, i.e., vastly different CNNs and ViTs. An overview of the T-TAME method is given in Fig. \ref{fig:overview}.

\subsection{T-TAME Overall Architecture}

The T-TAME method, as illustrated in Fig. \ref{fig:general}(a), is composed of the following components:
\begin{itemize}
    \item A feature map adapter
    \item The feature branches of the Attention Mechanism
    \item The fusion module of the Attention Mechanism
\end{itemize}
These components are trained, as illustrated in Fig. \ref{fig:overview}, using a suitable loss function, together with a mask selection and an image masking procedure. 

The feature map adapter reshapes the feature map set output by the backbone network so that it can be input to the attention mechanism, which consists of the feature branches and the fusion module. Each feature branch has a one-to-one mapping with each feature map in the feature map set and processes them separately. The fusion module combines the attention maps from each feature branch into the final class-discriminate explanation maps. Specific masks are then selected, in an unsupervised manner, and used to mask the image. The loss function takes as input a subset of the produced explanation maps, i.e., a number of slices along the channel dimension, and the logits generated by passing the masked image through the backbone network. In the next section, we specify each of these components.

\subsection{T-TAME Architecture Components}

\subsubsection{Attention Mechanism}
\label{sssec:AM}
For a feature map set $\boldsymbol{L}^s$, the attention mechanism consists of $s$ feature branches and the fusion module.
The feature branch structure consists of a $1 \times 1$ convolution layer with the same number of input and output channels, a batch normalization layer, a skip connection, and a ReLU activation, as illustrated in Fig. \ref{fig:general}(b).
Each feature branch
\begin{equation}
    FB: Sp\left(\boldsymbol{L}_i\right) \to Sp\left(\boldsymbol{A}_i\right)
\end{equation}
takes as input a single CNN-type feature map $\boldsymbol{L}_i$ (as defined in Eq.~\eqref{eq:cnn-fm}) and outputs an attention map 
\begin{equation}
    \boldsymbol{A}_i : \{1, \dotsc, C_i\} \times\{1, \dotsc, W_e\} \times \{1, \dotsc, H_e\} \to \mathbb{R},
\end{equation}
where $W_e = \max_i W_i$ and $H_e = \max_i H_i$.
That is, the attention map $\boldsymbol{A}_i$ has the same channel dimension as $\boldsymbol{L}_i$, and the same spatial dimensions as the explanation maps $\boldsymbol{E}$ (Eq.~\eqref{eq:expl-maps}). The dimensions $W_e$, $H_e$ are equal to the spatial dimensions of the largest input feature map. This is achieved by applying bilinear interpolation where necessary (Fig. \ref{fig:general}(b)), i.e., on the feature branches whose input feature map dimensions are smaller than $W_e$ and $H_e$.
The resulting attention maps $\boldsymbol{A}^s = \left\{\boldsymbol{A}_i \mid i \in \{1,\dotsc,s\}\right\}$ are forwarded into the fusion module
\begin{equation}
    FS: Sp\left(\boldsymbol{A}^s\right) \to Sp\left(\boldsymbol{E}\right),
\end{equation}
consisting of a concatenation operator, a $1 \times 1$ convolutional layer, and a sigmoid activation, as illustrated in Fig. \ref{fig:general}(c). Specifically, the attention maps are initially concatenated into a single attention map (a 3D tensor with $\sum_{i=1}^{s}C_i$ channels, each channel of spatial dimensions $W_e$, $H_e$), and then processed (by the convolution and sigmoid layers) to generate the explanation map.

\subsubsection{Feature Map Adapter}

In the context of a CNN backbone network, the feature maps inherently conform to the required input shape \((C_i,\, W_i,\, H_i)\) (as seen in Fig. \ref{fig:general}(b)), thus there is no need to adapt the feature maps to the attention mechanism.
In this case, the feature map adapter is the identity function $a(\boldsymbol{L}_i) = \boldsymbol{L}_i$.
When the backbone network is Transformer-based, as in the case of ViTs, the feature maps are defined as in Eq.~\eqref{eq:vitFm}.
The feature map adapter first excludes the class token, as it lacks spatial information, and then reshapes the feature map into a 3D format that mirrors the structure of feature maps typically found in a CNN backbone, as defined in Eq.~\eqref{eq:cnn-fm},
where $C_i = D$, $W_i=H_i=\sqrt{N}$.
This is essentially the inverse of the ViT architecture input processing step\footnote{In ViT, the feature map produced by the initial convolution layer, with dimensions 
$\left(D,\, \sqrt{N},\, \sqrt{N}\right)$ is initially reshaped into a 2D format with dimensions $\left(D,\, N\right)$.
Then, the order of dimensions is permuted, i.e., the dimensions become $\left(N,\, D\right)$ and the class token is introduced, resulting in a feature map with dimensions $\left(N+1,\, D\right)$.}. 

\subsubsection{Loss function, mask selection, and masking method}
\label{sssec:norm}

The loss function used for training the proposed attention mechanism is the weighted sum of two loss functions,
\begin{align}
\begin{split}
    Loss(\boldsymbol{\Psi},\ \text{logits},\ y)\ = \ &\lambda_1CE(\text{logits}, y) \\
    &+ \, \lambda_2TV'(\boldsymbol{\Psi}), 
\label{eq:loss}
\end{split}
\end{align}
where $CE()$, $TV'()$ are the cross-entropy and modified total variation loss, respectively; $\lambda_1$, $\lambda_2$ are the corresponding summation weights; and y is the predicted class of the backbone network (a.k.a. model truth): $y = \argmax f(\boldsymbol{I})$. $\boldsymbol{\Psi}$ is defined as
\begin{equation}
    \boldsymbol{\Psi} \colon  \lbrace\boldsymbol{E}_n \mid n \in \boldsymbol{Cls_{\boldsymbol{\Psi}}} \subset \lbrace1, \dotsc, Cls\rbrace\rbrace,
\end{equation}
i.e., $\boldsymbol{\Psi}$ is a set containing any number of explanation maps $\boldsymbol{E}_n$.
For each input image, in a batched training scenario with batch size $B$, we include in $\boldsymbol{\Psi}$ the explanation map corresponding to the predicted class $y$ of the backbone network, $\boldsymbol{E}_{\text{y}}$, and additional $B - 1$ explanation maps for randomly selected classes. The incorporation of explanation maps corresponding to other classes besides the predicted class in the loss function helps the attention mechanism to learn to generate class-discriminative explanation maps.

The cross-entropy loss uses the logits generated by the backbone network for the masked input image and the predicted class to compute a loss value. This term trains the attention mechanism to focus on salient, class-relevant parts of the input image.
The masking procedure involves taking the element-wise product (also known as the Hadamard product), denoted as $\odot$, between the raw image and the mask of the predicted class using,
\begin{eqnarray}
    \text{CNN Masking}(\boldsymbol{E}_y, \boldsymbol{I}) &=& \left|\boldsymbol{E}_y\odot \boldsymbol{I}\right|, \label{e:cnnmasking}\\
    \text{ViT Masking}(\boldsymbol{E}_y, \boldsymbol{I}) &=& \boldsymbol{E}_y \odot \left|\boldsymbol{I}\right|,  \label{e:vitmasking} 
\end{eqnarray}
where $\left| ~ \right|$ denotes element-wise standardization (also known as Z-score normalization) using the dataset mean and standard deviation \cite{standardization}. This operation shifts and scales each element of the input tensor based on the mean and standard deviation of the dataset.
We should note that masking removes features from the input image and renders it out-of-distribution \cite{JainICLR22}.
CNNs are more sensitive to such a transformation in comparison to Transformer-like architectures, as shown in \cite{vitrobust}.
To this end, in the case of CNN, the explanation map is first used as a mask to perturb the input image and then the standardization is applied (Eq.~\eqref{e:cnnmasking}).
This is the typical order of applying a perturbation (e.g. masking, augmentation, multiplicative/additive noise) in an input image, with the aim of causing a minimal shift to the input data distribution \cite{ChenArxiv20}.
On the other hand, in the case of ViT backbones, the image $\boldsymbol{I}$ is first standardized, and then used in the Hadamard product (Eq.~\eqref{e:vitmasking}).
This different approach is shown to perform better (see Table~\ref{tab:norm} in the Experiments section), and is motivated by considering what happens when standardizing only the input image: the explanation map, when used as a mask, behaves as a local perturbation, i.e., certain regions of the input image remain intact while the global statistics of the image change.
Since ViT-like models \cite{VaswaniNIPS2017,vit,JainICLR22} focus on certain image sub-regions and also examine global information, this type of perturbation is beneficial \cite{vitrobust,Zhang_2022_CVPR}.

The modified total variation loss, inspired by total variation denoising \cite{tvloss}, is the sum of the squares of the total variation norm of the explanation maps $\boldsymbol{\Psi}$ and the mean of element-wise exponentiation of the explanation maps. 
This term reduces noise and overactivation in the generated explanation maps. 
The modified total variation loss is defined as,
\begin{equation}
\label{eq:tvloss}
    TV'(\boldsymbol{\Psi}) = E(\boldsymbol{\Psi}) + \lambda_3V(\boldsymbol{\Psi}),
\end{equation}
with $E()$ defined as
\begin{equation}
\label{eq:arealoss}
E(\boldsymbol{\Psi}) = \frac{1}{S}\sum_{n,\, j,\,k} \boldsymbol{E}_{n,\, j,\,k}^{\lambda_4},\, \boldsymbol{E}_n \in \boldsymbol{\Psi},
\end{equation}
and $V()$ defined as
\begin{gather}
\begin{split}
    V(\boldsymbol{\Psi}) =\frac{1}{2 S} \sum_{n,\,j,\,k} \big( &\lvert\boldsymbol{E}_{n,\,j+1,\,k} - \boldsymbol{E}_{n,\,j,\,k} \rvert^2 + \\ &\lvert\boldsymbol{E}_{n,\,j,\,k+1} - \boldsymbol{E}_{n,\,j,\,k} \rvert^2 \big),\,
\boldsymbol{E}_n \in \boldsymbol{\Psi},
\end{split}
\end{gather}
where $\boldsymbol{E}_{n,\, j,\, k}$ denotes the value of the explanation map $\boldsymbol{E}_n$ in indices $\left(j, k\right)$ and $S = B \cdot W_e \cdot H_e$ is the number of such values included in the summation of Eq.~\eqref{eq:arealoss}.
$TV'(\boldsymbol{\Psi})$ forces the attention mechanism to output less noisy explanation maps that emphasize smaller and more focused regions in the input image instead of arbitrarily large areas. Without this term in the loss function, the trivial solution for minimizing the cross-entropy loss would be not masking the input image at all, with a homogeneous and appropriately scaled explanation map. The scalars $\lambda_3$ and $\lambda_4$ are additional hyperparameters of the loss function. By modifying the original total variation loss with the addition of these hyperparameters, we gain an additional degree of freedom to generate smoother and more focused explanation maps.

\subsection{Training \& Inference}

During the training of T-TAME, batches from the dataset that was used to originally train the backbone network are used to generate feature map sets and logits. The feature maps are then input to the attention mechanism to produce explanation maps. Using the predicted classes from the backbone's logits, specific explanation maps are selected and used to mask the input images. The batch of masked images is input to the backbone to produce new logits. The new logits and a subset of explanation maps corresponding to the predicted classes, as well as other random classes, are input to the loss function. Through backpropagation, the weights of the attention mechanism are optimized to produce more salient explanation maps. 

During inference, only the upper half of the architecture illustrated in Fig.~\ref{fig:general} is used: as typically done for classifying an input image, the image is input to the backbone classifier to generate a decision and, as an intermediate result of this process, a feature map set. Then, the produced feature map set is input to the trained attention mechanism for generating explanation maps for all classes of the backbone classifier.

We should clarify here that, at the inference stage, the sigmoid activation function of Fig. \ref{fig:general}(c) is replaced by a min-max scaling step. This is done to produce a heatmap in the $[0, 1]$ range, for a fair comparison with all of the examined explainability methods that typically introduce such a scaling step, e.g., \cite{selvaraju2017grad, chattopadhay2018grad, wang2020score}. Contrarily, the sigmoid function illustrated in Fig.~\ref{fig:general}(c) is used during training, because the gradient of the min-max scaling operation is very noisy, thus would impede training.

\section{Experiments}\label{sec:exp}

\subsection{Datasets and Backbone Networks}\label{ssec:dataset}

We choose three neural network models that are widely used for image classification, as the backbones for which we will generate explanations using T-TAME: VGG-16 \cite{simonyan2014very}, ResNet-50 \cite{he2016deep} and ViT-B-16 \cite{vit}. 
This choice is further motivated by the diversity among these models: there are significant differences between the two chosen CNN architectures, and between them and the ViT architecture. All 3 backbones have been trained on the ImageNet dataset \cite{imagenet-dataset};
we obtain the trained models from the \verb|torchvision.models| library.

For training and evaluating T-TAME on each of these backbones, we use the ImageNet ILSVRC 2012 dataset \cite{imagenet-dataset} (i.e., the same dataset that the backbones have been trained on).
This dataset contains 1000 classes, 1.3 million, and 50k images for training and evaluation, respectively.
Out of the last 50k images, we use a set of 2000 randomly selected images as the validation set and a different, disjoint set of 2000 randomly selected evaluation images for testing the explainability results (the same as in \cite{Gkartzonika2022, TAME} to allow for a fair comparison). The validation set is utilized for optimizing the T-TAME training hyperparameters, including the hyperparameters of the loss function: $\lambda_1$, $\lambda_2$, $\lambda_3$, and $\lambda_4$, as well as the number of training epochs and learning rate. Testing on 2000 images is chosen not only for consistency with \cite{Gkartzonika2022, TAME} but additionally because executing the perturbation-based approaches that we use in the experimental comparisons is computationally expensive \cite{wang2020score, petsiuk2018rise} (up to almost four orders of magnitude more expensive than T-TAME and gradient-based methods).   

\subsection{Evaluation measures}
\label{ss:eval_measures}

For quantitative evaluation and comparisons, we employ two widely used evaluation measures, Increase in Confidence (IC) and Average Drop (AD) \cite{chattopadhay2018grad}.
Additionally, we employ the promising Noisy Imputation method from the Remove and Debias (ROAD) evaluation framework recently introduced in  \cite{ROAD}.
For completeness, we briefly describe these two evaluation approaches in the following.

\subsubsection{IC and AD}

These two measures are defined as follows: 
\begin{align}
     \text{AD}(v) &= \sum_{\upsilon=1}^{\Upsilon}\frac{\max\lbrace 0, \psi_\upsilon - \psi^{\phi_v}_\upsilon\rbrace}{
     \Upsilon \psi_\upsilon} \cdot 100, \\
     \text{IC}(v) &= \sum_{\upsilon=1}^{\Upsilon}\frac{\text{int}\left(\psi^{\phi_v}_\upsilon>\psi_\upsilon\right)}{\Upsilon} \cdot 100,
\end{align}
where $\Upsilon$ represents the number of test images; $y_\upsilon = \argmax f(\boldsymbol{I}_\upsilon)$ is the model-truth label for the $\upsilon$th test image 
$\boldsymbol{I}_\upsilon$; and $\psi_\upsilon = \max f(\boldsymbol{I}_\upsilon)$ is the classifier's output score (confidence) for the model-truth class. $\psi^{\phi_v}_\upsilon$ is the classifier's output score for the model-truth class when input to the classifier is a modified image, i.e. one that is masked according to the explanation map for the same class, $\boldsymbol{E}_{y_\upsilon}$ (generated by the explainability method under evaluation). That is,  
\begin{gather}
    \psi^{\phi_v}_\upsilon = \boldsymbol{e}_{y_\upsilon}\cdot f\left(\boldsymbol{I}_\upsilon\odot \phi_v\left(\boldsymbol{E}_{y_\upsilon}\right)\right),\label{eq:maskscore}\\
    \boldsymbol{e}_{y_\upsilon} = (0, \dotsc, 1\text{ at position } y_\upsilon, \ldots, 0), \label{eq:basisvec}
\end{gather}
where $\phi_v()$ represents a threshold function to select the top $v\%$ higher-valued pixels of the explanation map $\boldsymbol{E}_{y_\upsilon}$, and $\text{int}()$ returns 1 when the input condition is satisfied and 0 otherwise. 

Intuitively, AD measures how much, on average, the produced explanation maps, when used to mask the input images, reduce the confidence of the model. The implicit assumption is that by masking the input image using the explanation, confusing and irrelevant background information is removed, and thus, the average drop in confidence should be minimized. In contrast, IC measures how often explanation maps, when applied in the same manner, increase the model's confidence. By eliminating confounding background information, 
the classification confidence likely will
increase, hence IC should be maximized. A naive all-ones mask would result in a $0\%$ AD, the optimal result, and $0\%$ IC, the worst result. Therefore, for a more comprehensive evaluation, we use the combination of these measures. Furthermore, since the explanation maps produced by each method vary in their intensity of activation, we also apply a threshold $v\%$ to the explanation maps, as discussed above, to assess how effectively the pixels are ordered based on importance. Using a smaller threshold (e.g. $v=15\%$) creates a more challenging evaluation setup since a smaller percentage of the image pixels is retained. This way, we can compare methods more fairly, since methods that produce highly activated explanation maps could initially generate good results without thresholding, but when a threshold is applied they may struggle, revealing a subpar ordering of pixel importance in the explanation map. This evaluation protocol has been adopted in most previous works, including \cite{petsiuk2018rise, wang2020score, ablationcam, chattopadhay2018grad, barkan2023visual, Gkartzonika2022, Chefer_2021_CVPR}.

\subsubsection{ROAD}

\begin{figure*}
     \centering
     \begin{subfigure}[t]{0.22\textwidth}
         \centering
         \includegraphics[width=\textwidth]{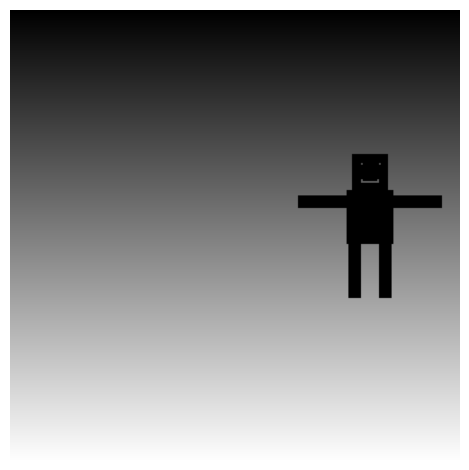}
         \caption{The original synthetic image.}
         \label{sfig:original}
     \end{subfigure}
     \hfill
     \begin{subfigure}[t]{0.22\textwidth}
         \centering
         \includegraphics[width=\textwidth]{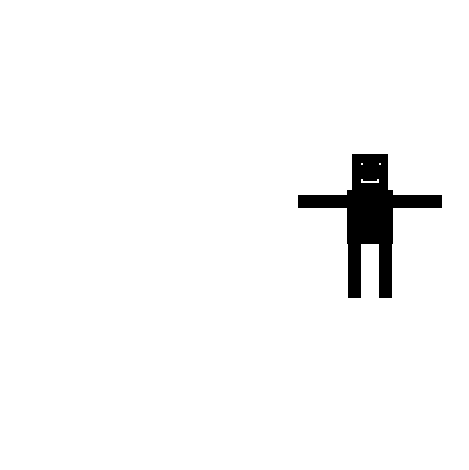}
         \caption{The binary mask of the pixels that will be imputed.}
         \label{sfig:binmask}
     \end{subfigure}
     \hfill
     \begin{subfigure}[t]{0.22\textwidth}
         \centering
         \includegraphics[width=\textwidth]{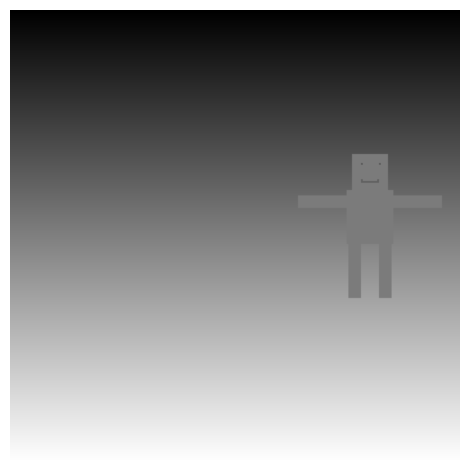}
         \caption{A naive imputation: the pixels indicated by the binary mask are replaced by the average pixel value (per channel) of the input image.}
         \label{sfig:naive}
     \end{subfigure}
     \hfill
    \begin{subfigure}[t]{0.22\textwidth}
         \centering
         \includegraphics[width=\textwidth]{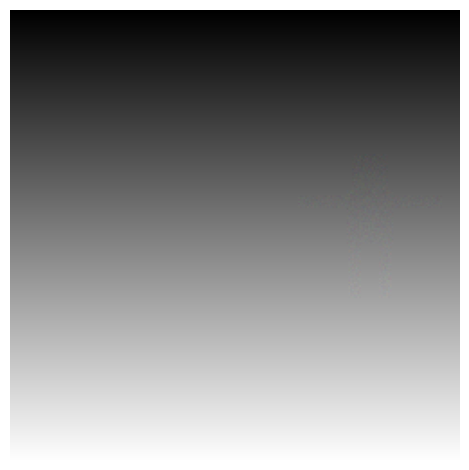}
         \caption{The imputation of the image using ROAD. In this case, it is much more difficult to discern which pixels were removed.}
         \label{sfig:roadmask}
     \end{subfigure}
        \caption{In this synthetic example, a typical imputation approach is compared to the noisy imputation method of the ROAD framework. In the naive case, information about the mask's shape is clearly leaked. Fig~\ref{sfig:roadmask} shows how ROAD removes pixels in a more nuanced way to avoid revealing the shape of the binary mask.}
        \label{fig:roads}
\end{figure*}

The Remove and Debias evaluation framework \cite{ROAD} aims to improve the process of assessing the quality of explanation maps of different explainability techniques with pixel perturbations.
The authors of \cite{ROAD} first prove, using an information theory analysis, that simpler methods of removing areas of an image using a binary mask leak information about the shape of the mask. The shape of the mask could reveal class information. Thus, the ROAD framework aims to remove salient information rather than simply removing salient pixels. An example of the effect of different imputation methods is shown in Fig.~\ref{fig:roads}.
In this example, we observe that by imputing the images in a straightforward way (Fig. \ref{sfig:naive}), i.e., replacing the removed pixels with the mean of the original image, the region of the modification of the original image is evident. This can leak information about the class contained in the image. Resolving the discrepancy between the removal of pixels and the removal of the information contained in the removed pixels is the aim of the noisy imputation method of ROAD. In Fig. \ref{sfig:roadmask}, where the noisy imputation method is employed, we observe that it is now much harder to detect which pixels were removed, reducing the leakage of class information contained in the binary mask.

Two evaluation measures are defined in ROAD, namely, MoRF (Most Relevant First) and LeRF (Least Relevant First).
In the former/latter, a binary mask generated from the explanation map is used that highlights the $v\%$ most/least important regions in the image.
This binary mask is then utilized to impute the input image, and the target logit, or the confidence in the target class, is calculated.
The ROAD score is then computed using,
\begin{align}
     \text{MoRF}(v) &= \sum_{\upsilon=1}^{\Upsilon}\frac{\psi^{\hat{\theta}_v}_\upsilon}{\Upsilon} \cdot 100, \\
     \text{LeRF}(v) &= \sum_{\upsilon=1}^{\Upsilon}\frac{\psi^{\check{\theta}_v}_\upsilon}{\Upsilon} \cdot 100,
\end{align}
where $\psi^{\hat{\theta}_v}_\upsilon = \boldsymbol{e}_{y_\upsilon} \cdot f(\hat{\theta}_v(\boldsymbol{I}_\upsilon, \boldsymbol{E}_{y_\upsilon})))$,
$\psi^{\check{\theta}_v}_\upsilon = \boldsymbol{e}_{y_\upsilon} \cdot f(\check{\theta}_v(\boldsymbol{I}_\upsilon, \boldsymbol{E}_{y_\upsilon})))$, $\boldsymbol{e}_{y_\upsilon}$ is defined in Eq.~\eqref{eq:basisvec}, and $\hat{\theta}_v()$, $\check{\theta}_v()$ represent the ROAD imputation operation applied to $v\%$ of the most or least important pixels of the input image, respectively.
In the case of MoRF, a sharp decline in model confidence should be observed, as the removal of important class information should rapidly deteriorate the model's performance.
In the case of LeRF, removing 
irrelevant information should minimally affect the confidence of the model.
We compute the ROAD measures only when comparing with other methods (i.e., in Section\ref{ss:benchmarking}), as ROAD is significantly more computationally expensive than computing the AD and IC measures. 
This new evaluation protocol has been adopted in the very recent works \cite{roaduse1, roaduse2}.

\subsection{Experimental setup}

\begin{figure*}[ht!]
\centering
\includegraphics[width=0.7\textwidth]{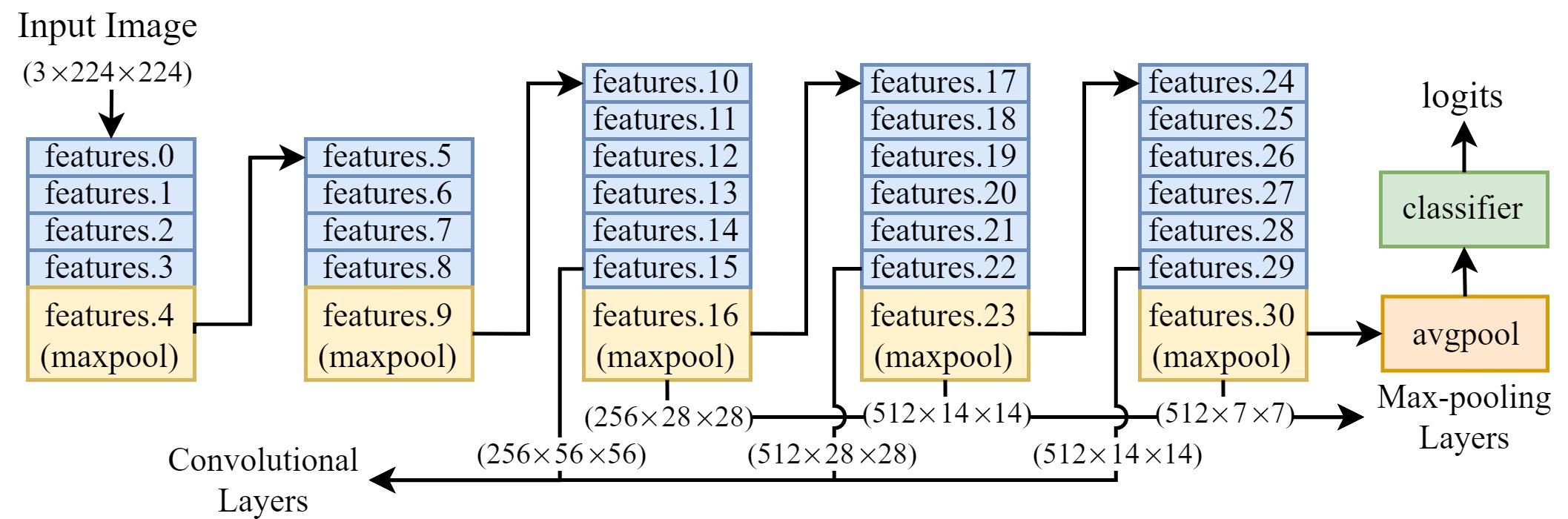} % exp.png
\caption{The layers from which feature maps are extracted when applying T-TAME to a VGG-16 backbone. We also indicate in this diagram the dimensions of the extracted feature maps. We experiment with two separate sets of layers in the ablation study (Table \ref{tab:abl}), where we denote by ``Max-pooling Layers'' the last three max-pooling layers, and by ``Convolutional Layers'' the three layers before the last three max-pooling layers. We use the same layer naming as the \texttt{torchvision.models.feature\_extraction} library.} 
\label{fig:vgg}
\end{figure*}
\begin{figure*}[ht!]
\centering
\includegraphics[width=0.6\textwidth]{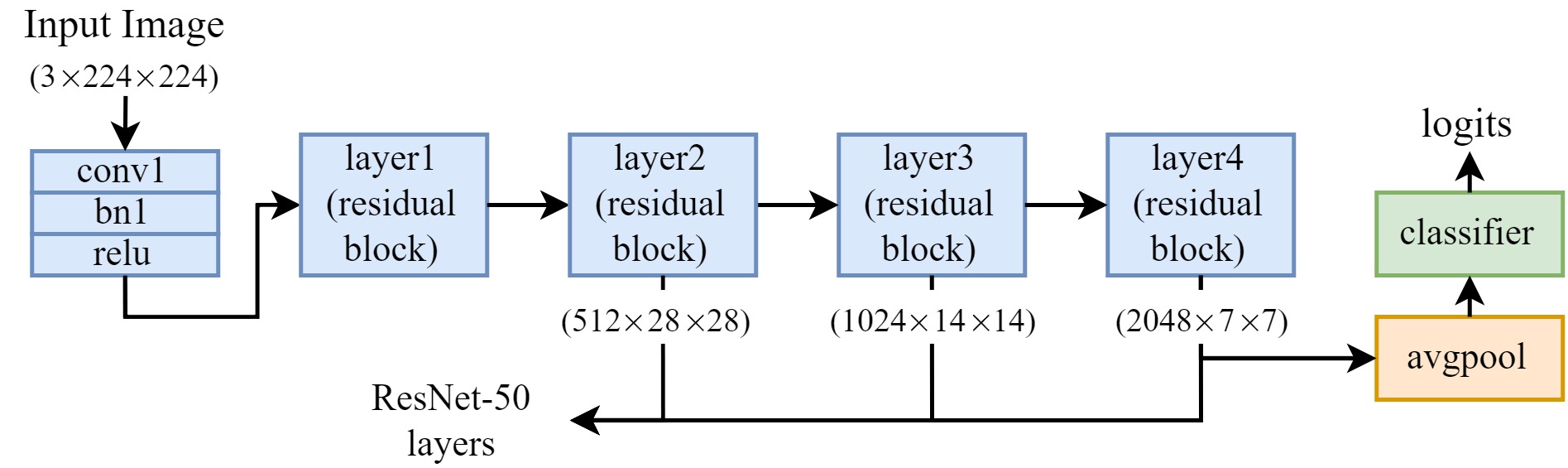} % exp.png
\caption{The layers from which feature maps are extracted when applying T-TAME to a ResNet-50 backbone. We also indicate in this diagram the dimensions of the extracted feature maps. The outputs of the final three residual blocks are used. We use the same layer naming as the \texttt{torchvision.models.feature\_extraction} library.}
\label{fig:resnet}
\end{figure*}
\begin{figure*}[ht!]
\centering
\includegraphics[width=0.6\textwidth]{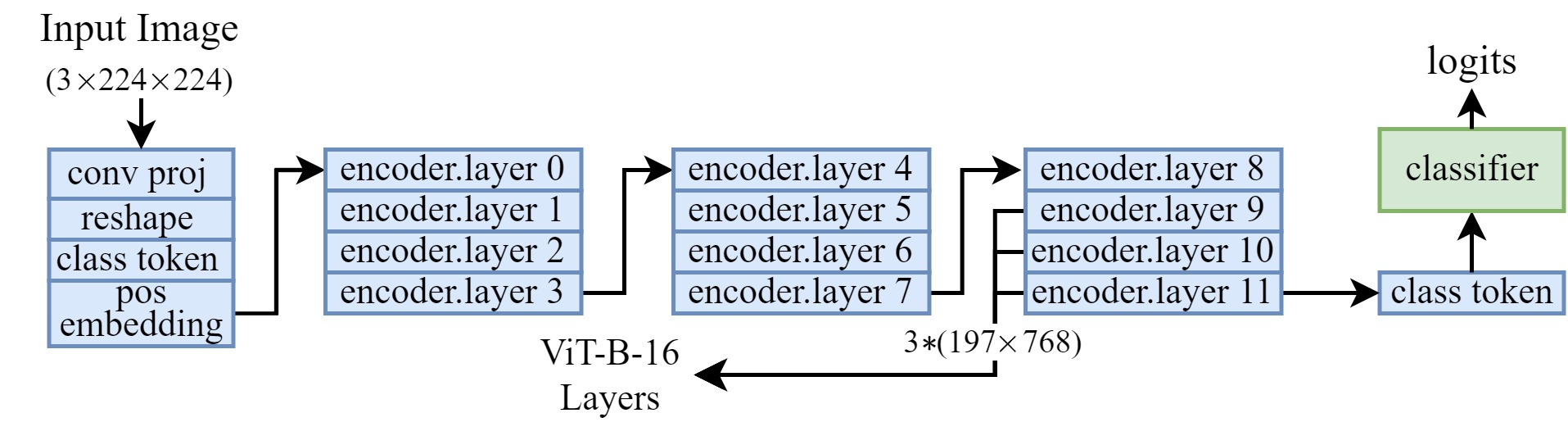} % exp.png
\caption{The layers from which feature maps are extracted when applying T-TAME to a ViT-B-16 backbone. We also indicate in this diagram the dimensions of the extracted feature maps. The outputs of the final three encoder blocks are used. We use the same layer naming as the \texttt{torchvision.models.feature\_extraction} library.}
\label{fig:vit}
\end{figure*}

Feature maps from three layers are extracted from each backbone to which T-TAME is applied (i.e., $s=3$).
The VGG-16 backbone model consists of five blocks of convolutions separated by $2\times 2$ max-pooling operations, as shown in Fig.~\ref{fig:vgg}.
We choose one layer from each of the last three blocks, namely the feature maps output by the max-pooling layers of each block.
Alternatively, we also experiment with the use of feature maps output by the last convolution layer of each block. The results of this alternate choice of feature maps are discussed in Section~\ref{sssec:abl}.
ResNet-50 consists of five stages, as depicted in Fig.~\ref{fig:resnet}.
For this backbone, we utilize the feature maps from its final three stages.
Finally, for the ViT-B-16 backbone, which consists of eleven encoder blocks, we use the feature maps of the last three encoder blocks, as shown in Fig.~\ref{fig:vit}.

T-TAME is trained using the loss function defined in Eq.~\eqref{eq:loss} with the SGD (Stochastic Gradient Descent) algorithm.
The OneCycleLR policy \cite{smith2019super} was utilized to vary the learning rate during the training procedure.
The largest batch size that can fit in the employed GPU's memory is used, as recommended in \cite{smith2018disciplined}. 
The rest of the hyperparameters were identified using the validation dataset and the $\text{IC}(15\%)$ and $\text{AD}(15\%)$ measures. IC and AD were preferred over ROAD because they are simpler to interpret and much less computationally expensive; and, we opted for IC and AD at the $v=15\%$ threshold because they are the most challenging ones to improve upon and provide more focused explanation maps. To this end, the optimal hyperparameters of the loss function (Eq.~\eqref{eq:loss}, Eq.~\eqref{eq:arealoss}, Eq.~\eqref{eq:tvloss}) were empirically identified as: $\lambda_1 = 1.5, \lambda_2 = 2, \lambda_3 = 0.005, \lambda_4 = 0.3$.
We observed surprising robustness across the different architectures using the above set of hyperparameters. Thus, the hyperparameter values do not vary between backbones.
The maximum learning rate in the OneCycleLR policy was optimized using a grid search.
Finally, the number of epochs was identified by varying it from one to eight and selecting the optimal one.

During training, the same image preprocessing employed in the original backbone network \cite{simonyan2014very, he2016deep, vit} is used, i.e., the smallest spatial dimension of each image is resized to 256 pixels, the image is then random-cropped to dimensions $W = H = 224$, and standardized using the channel-wise statistics calculated on the ImageNet dataset ($\text{mean} = [0.485, 0.456, 0.406]$, $\text{std} = [0.229, 0.224, 0.225]$).
During the testing phase, the image is again resized so that the smallest spatial dimension becomes 256 pixels, however, center-cropping is used instead of random-cropping, again as in \cite{simonyan2014very, he2016deep, vit}.
Subsequently, the appropriate masking procedure is selected, depending on the type of backbone network, as discussed in Section~\ref{sssec:norm}. This protocol is used unaltered for every considered explainability method, to ensure a fair comparison.
Feature maps are extracted from the backbone networks using the \verb|torchvision.models.feature_extraction| library.

\begin{table*}[htbp]
  \centering
  \caption{Comparison of T-TAME with other methods using the AD and IC measures (CNN backbones).}
   \begin{tabular}{lrrrrrrrrrr}\toprule
Backbone &Measure &Grad-CAM &Grad-CAM++ &Score-CAM &Ablation-CAM &RISE &IIA &L-CAM-Img &T-TAME \\
&&\cite{selvaraju2017grad}&\cite{chattopadhay2018grad}&\cite{wang2020score}& \cite{ablationcam} & \cite{petsiuk2018rise} & \cite{barkan2023visual}&\cite{Gkartzonika2022}&\\\midrule
\multirow{7}{*}{VGG-16} &AD↓(100\%) &32.12\% &30.75\% &27.75\% &34.87\% &\textbf{8.74\%} &25.42\% &12.15\% &\ul{9.33\%} \\
&IC↑(100\%) &22.10\% &22.05\% &22.80\% &19.25\% &\textbf{51.30\%} &22.90\% &40.75\% &\ul{50.00\%} \\
\cmidrule{2-10}
&AD↓(50\%) &58.65\% &54.11\% &45.60\% &49.23\% &42.42\% &64.43\% &\ul{37.37\%} &\textbf{36.50\%} \\
&IC↑(50\%) &9.50\% &11.15\% &14.10\% &11.45\% &17.55\% &5.60\% &\ul{20.25\%} &\textbf{22.45\%} \\
\cmidrule{2-10}
&AD↓(15\%) &84.15\% &82.72\% &75.70\% &76.96\% &78.70\% &87.68\% &\ul{74.23\%} &\textbf{73.29\%} \\
&IC↑(15\%) &2.20\% &3.15\% &4.30\% &36.50\% &\ul{4.45\%} &1.45\% &\ul{4.45\%} &\textbf{5.60\%} \\
\cmidrule{2-10}
&FP↓ &\textbf{1} &\textbf{1} &512 &2048 &4000 &\ul{12} &\textbf{1} &\textbf{1} \\
\midrule
\multirow{7}{*}{ResNet-50} &AD↓(100\%) &13.61\% &13.63\% &\ul{11.01\%} &13.58\% &11.12\% & 21.58\% &11.09\% &\textbf{7.81\%} \\
&IC↑(100\%) &38.10\% &37.95\% &39.55\% &37.05\% &\ul{46.15\%} &26.90\% &43.75\% &\textbf{54.00\%} \\
\cmidrule{2-10}
&AD↓(50\%) &29.28\% &30.37\% &\textbf{26.80\%} &30.33\% &36.31\% &41.31\% &29.12\% &\ul{27.88\%} \\
&IC↑(50\%) &23.05\% &23.45\% &\ul{24.75\%} &22.30\% &21.55\% &14.95\% &24.10\% &\textbf{27.50\%} \\
\cmidrule{2-10}
&AD↓(15\%) &\ul{78.61\%} &79.58\% &78.72\% &79.62\% &82.05\% &87.84\% &79.41\% &\textbf{78.58\%} \\
&IC↑(15\%) &3.40\% &3.40\% &3.60\% &3.50\% &3.20\% &1.45\% &\ul{3.90\%} &\textbf{4.90\%} \\
\cmidrule{2-10}
&FP↓ &\textbf{1} &\textbf{1} &2048 &8192 &8000 &\ul{12} &\textbf{1} &\textbf{1} \\
\bottomrule
\end{tabular}
  \label{tab:adic-cnn}%
\end{table*}%

\begin{table*}[htbp]
  \centering
  \caption{Comparison of T-TAME with other methods using the AD and IC measures (ViT backbone).}
   \begin{tabular}{lrrrrrrrrrr}\toprule
Backbone &Measure &Grad-CAM &Grad-CAM++ &Score-CAM &Ablation-CAM &RISE &Transformer LRP &IIA &T-TAME \\
&&\cite{selvaraju2017grad}&\cite{chattopadhay2018grad}&\cite{wang2020score}& \cite{ablationcam} & \cite{petsiuk2018rise} & \cite{Chefer_2021_CVPR}&\cite{barkan2023visual}&\\\midrule
\multirow{7}{*}{ViT-B-16} &AD↓(100\%) &37.19\% &57.21\% &38.09\% &35.50\% &\ul{13.52} &67.07\% &56.54\% &\textbf{8.19\%} \\
&IC↑(100\%) &12.75\% &5.55\% &15.35\% &8.90\% &\ul{37.00} &2.25\% &8.05\% &\textbf{38.35\%} \\
\cmidrule{2-10}
&AD↓(50\%) &40.74\% &72.77\% &44.20\% &42.16\% &\ul{31.94} &63.19\% &64.01\% &\textbf{23.64\%} \\
&IC↑(50\%) &12.30\% &4.85\% &14.50\% &10.55\% &\ul{33.10} &3.90\% &7.65\% &\textbf{40.40\%} \\
\cmidrule{2-10}
&AD↓(15\%) &\ul{73.11\%} &92.51\% &77.50\% &80.86\% &79.56 &88.68\% &86.44\%&\textbf{72.89\%} \\
&IC↑(15\%) &5.40\% &0.80\% &4.85\% &2.95\% &\ul{6.85} &0.90\% &4.00\% &\textbf{9.40\%} \\
\cmidrule{2-10}
&FP↓ &\textbf{1} &\textbf{1} &768 &768 &8000 &\textbf{1} &\ul{44} &\textbf{1} \\
\bottomrule
\end{tabular}
  \label{tab:adic-vit}%
\end{table*}%

\begin{table*}[htbp]
  \centering
  \caption{Comparison of T-TAME with other methods using the ROAD measures (CNN backbones).}
    \begin{tabular}{lrrrrrrrrrr}\toprule
Backbone &Measure &Grad-CAM &Grad-CAM++ &Score-CAM &Ablation-CAM &RISE &IIA &L-CAM-Img &T-TAME \\
&&\cite{selvaraju2017grad}&\cite{chattopadhay2018grad}&\cite{wang2020score}& \cite{ablationcam} & \cite{petsiuk2018rise} & \cite{barkan2023visual}&\cite{Gkartzonika2022}&\\\midrule
\multirow{2}{*}{VGG-16} &MoRF↓ &21.40\% &23.14\% &22.54\% &20.51\% &22.78\% &32.07\%&\ul{19.32\%}&\textbf{17.74\%} \\
&LeRF↑ &71.22\% &72.26\% &73.46\% &73.47\% &\textbf{76.08\%} &64.21\%&70.75\% &\ul{73.50\%} \\
\cmidrule{2-10}
&FP↓ &\textbf{1} &\textbf{1} &512 &2048 &4000 &\ul{12} &\textbf{1} &\textbf{1} \\
\midrule
\multirow{2}{*}{ResNet-50} &MoRF↓ &25.95\% &27.15\% &28.70\% &27.20\% &\textbf{23.80\%} &32.46\% &\ul{24.62\%} &25.83\% \\
&LeRF↑ &\ul{80.35\%} &79.22\% &78.50\% &79.16\% &\textbf{80.58\%} &73.95\% &76.69\%&75.35\% \\
\cmidrule{2-10}
&FP↓ &\textbf{1} &\textbf{1} &2048 &8192 &8000 &\ul{12} &\textbf{1} &\textbf{1} \\

\bottomrule
\end{tabular}
  \label{tab:road-cnn}%
\end{table*}%

\begin{table*}[htbp]
  \centering
  \caption{Comparison of T-TAME with other methods using the ROAD measures (ViT backbone).}
  \begin{tabular}{lrrrrrrrrrr}\toprule
Backbone &Measure &Grad-CAM &Grad-CAM++ &Score-CAM &Ablation-CAM &RISE &Transformer LRP &IIA &T-TAME \\
&&\cite{selvaraju2017grad}&\cite{chattopadhay2018grad}&\cite{wang2020score}& \cite{ablationcam} & \cite{petsiuk2018rise} & \cite{Chefer_2021_CVPR}&\cite{barkan2023visual}&\\\midrule
\multirow{2}{*}{ViT-B-16} &MoRF↓ &29.29\% &51.74\% &35.00\% &36.68\% &38.77\% &\ul{29.28\%} &42.37\% &\textbf{26.48\%} \\
&LeRF↑ &78.49\% &69.64\% &67.95\% &78.51\% &\textbf{82.81\%} &81.52\% &73.97\%&\ul{81.60\%} \\
\cmidrule{2-10}
&FP↓ &\textbf{1} &\textbf{1} &768 &768 &8000 &\textbf{1} &\ul{44} &\textbf{1} \\
\bottomrule
\end{tabular}
  \label{tab:road-vit}%
\end{table*}%

\begin{table*}[ht!]
\caption{Ablation study: different architectural choices of the attention mechanism of T-TAME.}
\label{tab:abl}
\centering
\begin{tabular}{ccccccccc}
\toprule
Model & Feature Extraction & Architecture Variant & $\text{AD↓}$ & $\text{IC↑}$ & $\text{AD↓}$ & $\text{IC↑}$ & $\text{AD↓}$ & $\text{IC↑}$ \\
&&&$(100\%)$&$(100\%)$&$(50\%)$&$(50\%)$&$(15\%)$&$(15\%)$\\
\midrule
\multirow{12}{*}{VGG-16} 
& \multirow{6}{*}{Max-pooling layers}
&T-TAME (Proposed) & 9.33 & 50 & 36.5 & 22.45 & \underline{73.29} & \underline{5.6} \\
&&No skip connection & 10.09 & 45.25 & 36.44 & 20.65 & 74.85 & 5.15 \\
&&No skip + No batch norm & \textbf{5.92} & \textbf {57.9} & \underline{34.49} & \textbf{24.2} & 74.58 & 5.15 \\
&&Sigmoid in feature branch & \underline{7.22} & \underline{55.65} & 38.4 & 21.6 & 79 & 4.85 \\
&&FMs from two layers & 10.72 & 45.45 & \textbf{34.48} & \underline{23.05} & \textbf{71.94} & \textbf{5.75} \\
&&FMs from ne layer & 12.1 & 42.1 & 35.81 & 20.8 & 74.19 & 4.85 \\
\cmidrule{2-9}
&\multirow{6}{*}{Convolutional layers}
&T-TAME (Proposed) & 9.07 & 51.1 & 40.72 & \textbf{20.9} & \underline{77.05} & \underline{4.8} \\
&&No skip connection & \textbf{6.22} & \underline{58.85} & 41.47 & \textbf{20.9} & 79.12 & 3.8 \\
&&No skip + No batch norm & \underline{6.62} & 56.6 & \textbf{40.48} & \underline{20.75} & 77.84 & \textbf{4.95} \\
&&Sigmoid in feature branch & 6.8 & \textbf{60} & 42.17 & 19.75 & 80.73 & 4.1 \\
&&FMs from two layers & 10.99 & 45.85 & \underline{40.89} & 19.55 & \textbf{76.66} & \underline{4.8} \\
&&FMs from one layer & 13.09 & 39.65 & 42.3 & 17.7 & 78.02 & 3.8 \\
\midrule
\multirow{6}{*}{ResNet-50} 
&\multirow{6}{*}{Stage Outputs}
&T-TAME (Proposed) & \underline{7.81} & \underline{54} & \underline{27.88} & \textbf{27.5} & \underline{78.58} & \textbf{4.9} \\
&&No skip connection & \textbf{5.7} & \textbf{62.65} & 46.58 & 18.25 & 89.32 & 2.3 \\
&&No skip + No batch norm & 9.29 & 50.25 & 29.43 & \underline{25.95} & 79.81 & 3.95 \\
&&Sigmoid in feature branch & 9.11 & 53.3 & 45.68 & 18.1 & 86.95 & 3.15 \\
&&FMs from two layers & 9.48 & 47.05 & \textbf{27.83} & 25 & \textbf{77.95} & \underline{4.25} \\
&&FMs from one layer & 11.32 & 43.45 & 29.85 & 24.25 & 79.59 & 3.55 \\
\midrule
\multirow{6}{*}{ViT-B-16} 
&\multirow{6}{*}{Block Outputs}
&T-TAME (Proposed) & \textbf{8.19} & \underline{38.35} & \textbf{23.64} &  \underline{40.40} & \textbf{72.89} & \textbf{9.40} \\
&&No skip connection & 10.16 & \textbf{38.65} & 24.42 & \textbf{41.65} & 73 & \underline{9.35} \\
&&No skip + No batch norm & \underline{8.91} & 36.75 & 24.79 & 39.75 & 72.98 & 8.70 \\
&&Sigmoid in feature branch & 12.75 & 35.50 & 26.79 & 38.85 & 72.99 & 8.85 \\
&&FMs from two layers & 9.23 & 37.70 & \underline{24.08} & 40 & 73.69 & 9.15 \\
&&FMs from one layer & 9.88 & 37.60 & 24.96 & 40.20 & \underline{72.97} & \textbf{9.40} \\
\bottomrule

\end{tabular}
\end{table*}

\begin{table}[htbp]
  \centering
  \caption{Ablation study: comparison of mismatched backbone-masking procedure combination.}
    \begin{tabular}{cp{4.215em}rr}
    \toprule
    \multicolumn{1}{p{4.215em}}{Backbone} & Measure & \multicolumn{1}{p{6.57em}}{ViT Masking (Eq.~\eqref{e:vitmasking})} & \multicolumn{1}{p{6.5em}}{CNN Masking (Eq.~\eqref{e:cnnmasking})} \\
    \midrule
    \multicolumn{1}{c}{\multirow{6}[6]{*}{VGG-16}} & AD↓(100\%) & 12.44 & \textbf{9.33} \\
      & IC↑(100\%) & 42.05 & \textbf{50.00} \\
\cmidrule{2-4}      & AD↓(50\%) & 41.38 & \textbf{36.50} \\
      & IC↑(50\%) & 18.05 & \textbf{22.45} \\
\cmidrule{2-4}      & AD↓(15\%) & 76.81 & \textbf{73.29} \\
      & IC↑(15\%) & 4.90 & \textbf{5.60} \\
    \midrule
    \multicolumn{1}{c}{\multirow{6}[6]{*}{ResNet-50}} & AD↓(100\%) & 16.97 & \textbf{7.81} \\
      & IC↑(100\%) & 37.20 & \textbf{54.00} \\
\cmidrule{2-4}      & AD↓(50\%) & 73.56 & \textbf{27.88} \\
      & IC↑(50\%) & 6.85 & \textbf{27.50} \\
\cmidrule{2-4}      & AD↓(15\%) & 96.10 & \textbf{78.58} \\
      & IC↑(15\%) & 0.90 & \textbf{4.90} \\
    \midrule
    
      \multicolumn{1}{c}{\multirow{6}[6]{*}{ViT-B-16}} & AD↓(100\%) & \textbf{8.19} & 9.76 \\
      & IC↑(100\%) & 38.35 & \textbf{39.45} \\
\cmidrule{2-4}      & AD↓(50\%) & \textbf{23.64} & 25.49 \\
      & IC↑(50\%) & 40.40 & \textbf{41.6} \\
\cmidrule{2-4}      & AD↓(15\%) & \textbf{72.89} & 74.85 \\
      & IC↑(15\%) & \textbf{9.40} & 8.20 \\
    \bottomrule
    \end{tabular}%
  \label{tab:norm}%
\end{table}%

\begin{figure}
\centering
\begin{tikzpicture}[scale=1]
\begin{axis}[
    legend style={nodes={scale=0.8, transform shape}},
    xlabel={\% Removed (MoRF↓)}, 
    ylabel={\% Average Confidence}, 
    grid
]
\addplot[
    color=cyan,
    mark=x,
    mark options={solid},
    dashed
    ]
    table[col sep=comma, x=Labels, y=vgg16 gradcam]{road_data.csv};
    \addlegendentry{GradCAM}
\addplot[
    color=green,
    mark=*,
    mark options={solid},
    dashed
    ]
    table[col sep=comma, x=Labels, y=vgg16 gradcam++]{road_data.csv};
    \addlegendentry{GradCAM++}
\addplot[
    color=red,
    mark=oplus,
    mark options={solid},
    dashed
    ] table[col sep=comma, x=Labels, y=vgg16 scorecam]{road_data.csv};
    \addlegendentry{ScoreCAM}
\addplot[
    color=violet,
    mark=+,
    mark options={solid},
    dashed
    ]
    table[col sep=comma, x=Labels, y=vgg16 ablationcam]{road_data.csv};
    \addlegendentry{AblationCAM}
\addplot[
    color=teal,
    mark=square*,
    mark options={solid},
    dashed
    ]
    table[col sep=comma, x=Labels, y=RISE vgg16]{road_data.csv};
    \addlegendentry{RISE}
\addplot[
    color=orange,
    mark=triangle*,
    mark options={solid},
    dashed
    ]
    table[col sep=comma, x=Labels, y=iia vgg16]{road_data.csv};
    \addlegendentry{IIA}
\addplot[
    color=olive,
    mark=diamond*,
    mark options={solid},
    dashed
    ]
    table[col sep=comma, x=Labels, y=lcam vgg16]{road_data.csv};
    \addlegendentry{L-CAM-Img}
\addplot[
    color=magenta,
    mark=otimes,
    mark options={solid},
    dashed
    ]
    table[col sep=comma, x=Labels, y=vgg16 TAME]{road_data.csv};
    \addlegendentry{T-TAME}

\end{axis}
\end{tikzpicture}
\caption{Comparison of methods using the MoRF measure of ROAD on the VGG-16 backbone.
}
\label{fig:morf-vgg}
\end{figure}
\begin{figure}
\centering
\begin{tikzpicture}[scale=1]
\begin{axis}[
    legend style={at={(0.02,0.02)},anchor=south west, nodes={scale=0.8, transform shape}},
    xlabel={\% Removed (LeRF↑)}, 
    ylabel={\% Average Confidence}, 
    grid
]
\addplot[
    color=cyan,
    mark=x,
    mark options={solid},
    dashed
    ]
    table[col sep=comma, x=Labels, y=vgg16 gradcam]{road_data_l.csv};
    \addlegendentry{GradCAM}
\addplot[
    color=green,
    mark=*,
    mark options={solid},
    dashed
    ]
    table[col sep=comma, x=Labels, y=vgg16 gradcam++]{road_data_l.csv};
    \addlegendentry{GradCAM++}
\addplot[
    color=red,
    mark=oplus,
    mark options={solid},
    dashed
    ] table[col sep=comma, x=Labels, y=vgg16 scorecam]{road_data_l.csv};
    \addlegendentry{ScoreCAM}
\addplot[
    color=violet,
    mark=+,
    mark options={solid},
    dashed
    ]
    table[col sep=comma, x=Labels, y=vgg16 ablationcam]{road_data_l.csv};
    \addlegendentry{AblationCAM}
\addplot[
    color=teal,
    mark=square*,
    mark options={solid},
    dashed
    ]
    table[col sep=comma, x=Labels, y=vgg16 RISE]{road_data_l.csv};
    \addlegendentry{RISE}
\addplot[
    color=orange,
    mark=triangle*,
    mark options={solid},
    dashed
    ]
    table[col sep=comma, x=Labels, y=iia vgg16]{road_data_l.csv};
    \addlegendentry{IIA}
\addplot[
    color=olive,
    mark=diamond*,
    mark options={solid},
    dashed
    ]
    table[col sep=comma, x=Labels, y=lcam vgg16]{road_data_l.csv};
    \addlegendentry{L-CAM-Img}
\addplot[
    color=magenta,
    mark=otimes,
    mark options={solid},
    dashed
    ]
    table[col sep=comma, x=Labels, y=vgg16 TAME]{road_data_l.csv};
    \addlegendentry{T-TAME}

\end{axis}
\end{tikzpicture}
\caption{Comparison of methods using the LeRF measure of ROAD on the VGG-16 backbone.
}
\label{fig:lerf-vgg}
\end{figure}

\begin{figure}
\centering
\begin{tikzpicture}[scale=1]
\begin{axis}[
    legend style={nodes={scale=0.8, transform shape}},
    xlabel={\% Removed (MoRF↓)}, 
    ylabel={\% Average Confidence}, 
    grid
]
\addplot[
    color=cyan,
    mark=x,
    mark options={solid},
    dashed
    ]
    table[col sep=comma, x=Labels, y=resnet50 gradcam]{road_data.csv};
    \addlegendentry{GradCAM}
\addplot[
    color=green,
    mark=*,
    mark options={solid},
    dashed
    ]
    table[col sep=comma, x=Labels, y=resnet50 gradcam++]{road_data.csv};
    \addlegendentry{GradCAM++}
\addplot[
    color=red,
    mark=oplus,
    mark options={solid},
    dashed
    ] table[col sep=comma, x=Labels, y=resnet50 scorecam]{road_data.csv};
    \addlegendentry{ScoreCAM}
\addplot[
    color=violet,
    mark=+,
    mark options={solid},
    dashed
    ]
    table[col sep=comma, x=Labels, y=resnet50 ablationcam]{road_data.csv};
    \addlegendentry{AblationCAM}
\addplot[
    color=teal,
    mark=square*,
    mark options={solid},
    dashed
    ]
    table[col sep=comma, x=Labels, y=RISE resnet50]{road_data.csv};
    \addlegendentry{RISE}
\addplot[
    color=orange,
    mark=triangle*,
    mark options={solid},
    dashed
    ]
    table[col sep=comma, x=Labels, y=iia resnet50]{road_data.csv};
    \addlegendentry{IIA}
\addplot[
    color=olive,
    mark=diamond*,
    mark options={solid},
    dashed
    ]
    table[col sep=comma, x=Labels, y=lcam resnet50]{road_data.csv};
    \addlegendentry{L-CAM-Img}
\addplot[
    color=magenta,
    mark=otimes,
    mark options={solid},
    dashed
    ]
    table[col sep=comma, x=Labels, y=resnet50 TAME]{road_data.csv};
    \addlegendentry{T-TAME}

\end{axis}
\end{tikzpicture}
\caption{Comparison of methods using the MoRF measure of ROAD on the ResNet-50 backbone.
}
\label{fig:morf-resnet}
\end{figure}
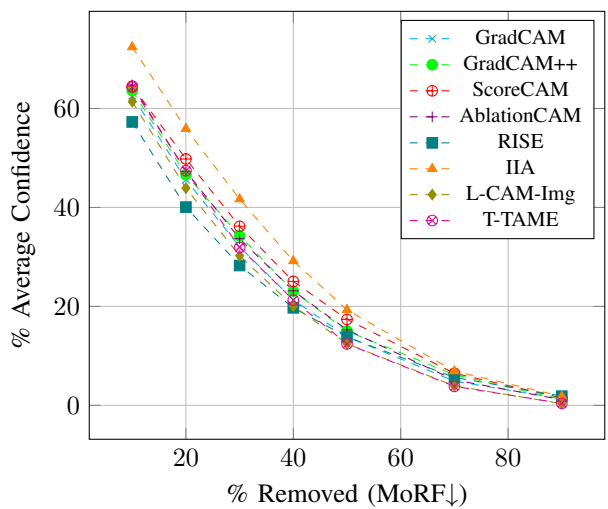
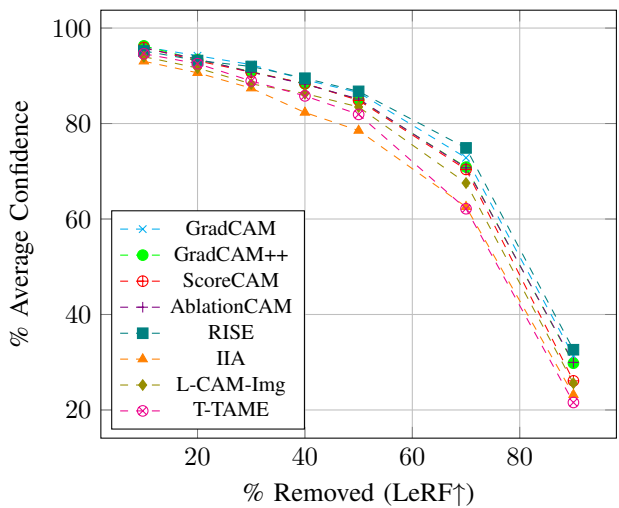
\begin{figure}
\centering
\begin{tikzpicture}[scale=1]
\begin{axis}[
    legend style={at={(0.02,0.02)},anchor=south west, nodes={scale=0.8, transform shape}},
    xlabel={\% Removed (LeRF↑)}, 
    ylabel={\% Average Confidence}, 
    grid
]
\addplot[
    color=cyan,
    mark=x,
    mark options={solid},
    dashed
    ]
    table[col sep=comma, x=Labels, y=resnet50 gradcam]{road_data_l.csv};
    \addlegendentry{GradCAM}
\addplot[
    color=green,
    mark=*,
    mark options={solid},
    dashed
    ]
    table[col sep=comma, x=Labels, y=resnet50 gradcam++]{road_data_l.csv};
    \addlegendentry{GradCAM++}
\addplot[
    color=red,
    mark=oplus,
    mark options={solid},
    dashed
    ] table[col sep=comma, x=Labels, y=resnet50 scorecam]{road_data_l.csv};
    \addlegendentry{ScoreCAM}
\addplot[
    color=violet,
    mark=+,
    mark options={solid},
    dashed
    ]
    table[col sep=comma, x=Labels, y=resnet50 ablationcam]{road_data_l.csv};
    \addlegendentry{AblationCAM}
\addplot[
    color=teal,
    mark=square*,
    mark options={solid},
    dashed
    ]
    table[col sep=comma, x=Labels, y=resnet50 RISE]{road_data_l.csv};
    \addlegendentry{RISE}
\addplot[
    color=orange,
    mark=triangle*,
    mark options={solid},
    dashed
    ]
    table[col sep=comma, x=Labels, y=iia resnet50]{road_data_l.csv};
    \addlegendentry{IIA}
\addplot[
    color=olive,
    mark=diamond*,
    mark options={solid},
    dashed
    ]
    table[col sep=comma, x=Labels, y=lcam resnet50]{road_data_l.csv};
    \addlegendentry{L-CAM-Img}
\addplot[
    color=magenta,
    mark=otimes,
    mark options={solid},
    dashed
    ]
    table[col sep=comma, x=Labels, y=resnet50 TAME]{road_data_l.csv};
    \addlegendentry{T-TAME}

\end{axis}
\end{tikzpicture}
\caption{Comparison of methods using the LeRF measure on the ResNet-50 backbone.
}
\label{fig:lerf-resnet}
\end{figure}
\begin{figure}
\centering
\begin{tikzpicture}[scale=1]
\begin{axis}[
    legend style={nodes={scale=0.8, transform shape}},
    xlabel={\% Removed (MoRF↓)}, 
    ylabel={\% Average Confidence}, 
    grid
]
\addplot[
    color=cyan,
    mark=x,
    mark options={solid},
    dashed
    ]
    table[col sep=comma, x=Labels, y=gradcam correct vit]{road_data.csv};
    \addlegendentry{GradCAM}
\addplot[
    color=green,
    mark=*,
    mark options={solid},
    dashed
    ]
    table[col sep=comma, x=Labels, y=gradcam++ correct vit]{road_data.csv};
    \addlegendentry{GradCAM++}
\addplot[
    color=red,
    mark=oplus,
    mark options={solid},
    dashed
    ] table[col sep=comma, x=Labels, y=scorecam correct vit]{road_data.csv};
    \addlegendentry{ScoreCAM}
\addplot[
    color=violet,
    mark=+,
    mark options={solid},
    dashed
    ]
    table[col sep=comma, x=Labels, y=ablationcam correct vit]{road_data.csv};
    \addlegendentry{AblationCAM}
\addplot[
    color=teal,
    mark=square*,
    mark options={solid},
    dashed
    ]
    table[col sep=comma, x=Labels, y=RISE vit_b_16]{road_data.csv};
    \addlegendentry{RISE}
\addplot[
    color=blue,
    mark=triangle*,
    mark options={solid},
    dashed
    ]
    table[col sep=comma, x=Labels, y=hila correct vit]{road_data.csv};
    \addlegendentry{Transformer LRP}
\addplot[
    color=orange,
    mark=diamond*,
    mark options={solid},
    dashed
    ]
    table[col sep=comma, x=Labels, y=iia vit_b_16]{road_data.csv};
    \addlegendentry{IIA}
\addplot[
    color=magenta,
    mark=otimes,
    mark options={solid},
    dashed
    ]
    table[col sep=comma, x=Labels, y=TAME vit_b_16 new]{road_data.csv};
    \addlegendentry{T-TAME}

\end{axis}
\end{tikzpicture}
\caption{Comparison of methods using the MoRF measure of ROAD on the ViT-B-16 backbone.
}
\label{fig:morf-vit}
\end{figure}
\begin{figure}
\centering
\begin{tikzpicture}[scale=1]
\begin{axis}[
    legend style={at={(0.02,0.02)},anchor=south west, nodes={scale=0.8, transform shape}},
    xlabel={\% Removed (LeRF↑)}, 
    ylabel={\% Average Confidence}, 
    grid
]
\addplot[
    color=cyan,
    mark=x,
    mark options={solid},
    dashed
    ]
    table[col sep=comma, x=Labels, y=vit_b_16 gradcam]{road_data_l.csv};
    \addlegendentry{GradCAM}
\addplot[
    color=green,
    mark=*,
    mark options={solid},
    dashed
    ]
    table[col sep=comma, x=Labels, y=vit_b_16 gradcam++]{road_data_l.csv};
    \addlegendentry{GradCAM++}
\addplot[
    color=red,
    mark=oplus,
    mark options={solid},
    dashed
    ] table[col sep=comma, x=Labels, y=vit_b_16 scorecam]{road_data_l.csv};
    \addlegendentry{ScoreCAM}
\addplot[
    color=violet,
    mark=+,
    mark options={solid},
    dashed
    ]
    table[col sep=comma, x=Labels, y=vit_b_16 ablationcam]{road_data_l.csv};
    \addlegendentry{AblationCAM}
\addplot[
    color=teal,
    mark=square*,
    mark options={solid},
    dashed
    ]
    table[col sep=comma, x=Labels, y=vit_b_16 RISE]{road_data_l.csv};
    \addlegendentry{RISE}
\addplot[
    color=blue,
    mark=triangle*,
    mark options={solid},
    dashed
    ]
    table[col sep=comma, x=Labels, y=vit_b_16 hila]{road_data_l.csv};
    \addlegendentry{Transformer LRP}
\addplot[
    color=orange,
    mark=diamond*,
    mark options={solid},
    dashed
    ]
    table[col sep=comma, x=Labels, y=iia vit_b_16]{road_data_l.csv};
    \addlegendentry{IIA}
\addplot[
    color=magenta,
    mark=otimes,
    mark options={solid},
    dashed
    ]
    table[col sep=comma, x=Labels, y=vit_b_16 TAME]{road_data_l.csv};
    \addlegendentry{T-TAME}

\end{axis}
\end{tikzpicture}
\caption{Comparison of methods using the LeRF measure of ROAD on the ViT-B-16 backbone.
}
\label{fig:lerf-vit}
\end{figure}

\subsection{Quantitative analysis}
\label{ss:benchmarking}

The following state-of-the-art methods are quantitatively compared with the proposed T-TAME, on all three considered backbones, using the evaluation measures described in Section~\ref{ss:eval_measures}: Grad-CAM \cite{selvaraju2017grad}, Grad-CAM++ \cite{chattopadhay2018grad}, Score-CAM \cite{wang2020score}, Ablation-CAM \cite{ablationcam}, RISE \cite{petsiuk2018rise} and Iterated Integrated Attributions (IIA) \cite{barkan2023visual}.
Additionally, we compare with L-CAM-Img \cite{Gkartzonika2022} and Transformer Layer-wise Relevance Propagation (LRP) \cite{Chefer_2021_CVPR}, only on CNN and Transformer backbones, respectively (because L-CAM-Img and Transformer LRP are only applicable to these specific backbones). These methods are selected because they are among the top-performing methods in the visual XAI domain and their source code is publicly available. For Grad-CAM \cite{selvaraju2017grad}, Grad-CAM++ \cite{chattopadhay2018grad}, Score-CAM \cite{wang2020score}, Ablation-CAM \cite{ablationcam}, we use the implementations of the \verb|pytorch_gradcam| library \cite{jacobgilpytorchcam}. For RISE \cite{petsiuk2018rise} and Iterated Integrated Attributions (IIA) \cite{barkan2023visual} we use the original implementations available at \url{https://github.com/eclique/RISE} and \url{https://github.com/iia-iccv23/iia}, respectively.
For L-CAM-Img \cite{Gkartzonika2022}, which is only applicable to CNN backbones, we use the original implementation, available at \url{https://github.com/bmezaris/L-CAM}. Finally, for the Transformer Layer-wise Relevance Propagation (LRP) method \cite{Chefer_2021_CVPR}, which is only applicable to ViT backbones, we use the original implementation, available at \url{https://github.com/hila-chefer/Transformer-Explainability}.

The results in terms of the $\text{AD}(v)$ and $\text{IC}(v)$ measures with $v=15\%, 50\%, 100\%$
for CNN and ViT models are shown in Tables \ref{tab:adic-cnn} and \ref{tab:adic-vit}.
The respective results for the
$\text{MoRF}(v)$ and $\text{LeRF}(v)$ measures are shown in Figs. \ref{fig:morf-vgg} to \ref{fig:lerf-vit}, where $v$ varies from $10\%$ to $90\%$.
In order to acquire a single value for each ROAD measure, model, and examined explainability method, we also compute the average confidence score across all percentages $v$ described above; these results are presented in Tables \ref{tab:road-cnn} and \ref{tab:road-vit}.

In all tables, for each comparison (i.e., each row), the best and second-best results are shown in bold and underline.
From the obtained results, we observe the following:
\begin{enumerate}[label=(\roman*)]

\item For the CNN backbones, T-TAME generally provides the best performance. Specifically, in the case of the VGG-16 backbone, for the AD and IC measures, T-TAME provides the best results for the more challenging $v=50\%$ and $v=15\%$ thresholds and is only outperformed in the less-challenging $v=100\%$ setup by the perturbation-based method RISE, which requires 4000 forward passes to generate a single explanation (thus, being 4000 times more computationally expensive than T-TAME at inference time). In the case of the ResNet-50 backbone, for the AD and IC measures, T-TAME is overall the top-performing method, while being second-best in one instance. In that instance, it is outperformed by the perturbation-based method Score-CAM, which however requires 2048 forward passes (instead of one, for T-TAME) to generate a single explanation.

From the averaged ROAD measures of Table~\ref{tab:road-cnn}, in the case of the VGG-16 backbone, we observe that T-TAME achieves the best results w.r.t. the MoRF measure. According to the LeRF measure, it is outperformed only by RISE, as in the case of the AD and IC metrics. 
In the case of the ResNet-50 backbone, from Table~\ref{tab:road-cnn} and Fig.~\ref{fig:morf-resnet} we observe that w.r.t. MoRF, the explanation maps of RISE produce the lowest (i.e., best) average confidence
but are overtaken by T-TAME in the higher removal percentages (50\% or more, in Fig.~\ref{fig:morf-resnet}). These results suggest that T-TAME correctly identifies the important regions, but the exact pixel-wise importance ordering is noisy. Additionally, w.r.t. LeRF for this backbone, RISE has the highest average confidence. To explain the LeRF results of T-TAME in the case of ResNet-50, we should recall that LeRF is computed by removing the less important features of the input image, and takes into account only the ordering of pixels according to the explanation map. As can be seen in Fig.~\ref{fig:backbones}, because of the low-resolution feature maps that are due to the specifics of the ResNet-50 architecture, the produced explanations are overly smooth. While they highlight the important regions of the input image, the ordering of less important pixels is noisy. Since for the computation of the ROAD measures the ordering of pixel importance is the only consideration for their computation, this quality is detrimental. Still, the ability of T-TAME to generate explanation maps in a single forward pass is a significant advantage for practical applications.

\item For the ViT-B-16 backbone, T-TAME is the top-performing method across the board. It performs best for all thresholds of the AD and IC measures. It is the second-best method only in the case of the LeRF measure, being outperformed by RISE, which, in this case, requires 8000 forward passes to generate a single explanation. Moreover, T-TAME outperforms the Transformer-specific LRP-based method. In the case of MoRF, as observed in Fig.~\ref{fig:morf-vit}, T-TAME exhibits the overall best performance for all percentages except for the initial $v=10\%$ removal percentage. Particularly for $v=30\%$ to $v=70\%$, the difference between T-TAME and the second-best method, Transformer LRP, is large. This suggests that, except for the very fine-grained ordering examined in the case of $v=10\%$, T-TAME correctly identifies the most important pixels for the ViT-B-16 backbone. In the case of LeRF, RISE is initially the top-performing method, being outperformed by T-TAME in the higher removal percentages. This again suggests a more globally correct ordering of importance, with less finely-grained orderings in the lower percentages. Considering that T-TAME requires only one forward pass to compute an explanation, it is significant that it can compete with and in most cases outperform perturbation-based approaches.
\end{enumerate}

\subsection{Ablation studies}
\label{ss:ablation}

In this section, we perform several ablation studies to assess the effects of different architectural choices of the T-TAME attention mechanism and to observe the effect of the different masking procedures when a CNN (Eq.~\eqref{e:cnnmasking}) or ViT backbone (Eq.~\eqref{e:vitmasking}) is used.
We measure performance utilizing only the $\text{AD}(v)$ and $\text{IC}(v)$ measures, to allow a more straightforward interpretation of the results, and additionally, because the ROAD measures are much more computationally expensive to compute.

\subsubsection{Different architectural choices of the attention mechanism} \label{sssec:abl}

Results of this set of ablation experiments are reported in Table~\ref{tab:abl}, where we indicate with bold/underline the best/second best results according to each measure for each model and layer selection.
For the VGG-16 model, inspired by similar works in the literature suggesting that the last layers of the network provide more salient features \cite{jetley2018learn}, we report two sets of experiments, one that uses features maps extracted from the three last max-pooling layers and one where feature maps are extracted from the layers directly before the last three max-pooling layers (Fig.~\ref{fig:vgg}). 
There is a difference in the spatial dimensions of the explanation maps generated using the former and the latter layers for feature extraction, i.e., $28 \times 28$ versus $56\times 56$, since the dimension of the explanation maps obtained by T-TAME is dictated by that of the employed feature maps (as explained in Section~\ref{sssec:AM}). For the ResNet-50 model, we extract feature maps from the outputs of the final three stages, resulting in an explanation map of $28 \times 28$ pixels. In the case of the ViT-B-16 model, feature maps are extracted from the outputs of the final three encoder blocks, resulting in an explanation map of $14 \times 14$ pixels. For each backbone and set of considered feature maps, we examine the following variants of the proposed architecture:

\textit{No skip connection}: It has been shown that the inclusion of a skip connection promotes a smoother loss landscape \cite{li2018visualizing} and preserves gradients that might otherwise be lost or diluted by passing through multiple layers, thus improving the training of very deep neural networks. Even for shallower neural networks, such as the proposed attention mechanism, we can benefit from using a skip connection. We see that by omitting the skip connection shown in Fig.~\ref{fig:general}(a), we get worse results in ResNet-50 for the more challenging $v=50\%$ and $v=15\%$ measures. Similarly, for the VGG-16 backbone, we report worse performance for the harder $v=50\%$ and $v=15\%$ measures. In the case of ViT-B-16, the proposed architecture that includes this skip connection prevails in the more challenging $v=15\%$ metric.

\textit{No skip + No batch norm}: Batch normalization is used in neural networks for speeding up training and combating internal covariate shift \cite{ioffe2015batch}.
Compared to the proposed architecture of Fig.~\ref{fig:general}(a), we see that, in the case of VGG-16, this variant generally performs better in the $v=100\%$ measures, but this does not hold for the other measures.

\textit{Sigmoid in feature branch}: In this variant, we replace the ReLU function of Fig.~\ref{fig:general}(a) with the sigmoid function, which squeezes the input from $(-\infty, \infty)$ to the output $(0, 1)$.
It is well known that the sigmoid function in deeper neural networks causes the vanishing gradient problem, making it more difficult to train the early layers of the neural network. We see again that the proposed architecture of Fig.~\ref{fig:general}(a) prevails for the more challenging $v=15\%$ measures.

\textit{Two layers} and \textit{One layer}: In this case, the proposed attention mechanism architecture is employed with feature maps from fewer than three layers.
The results when using just one layer, i.e., omitting the two earlier layers of the backbone (Fig.~\ref{fig:vgg}), are very similar to the L-CAM-Img method (as shown in Table~\ref{tab:adic-cnn}), which also uses just one feature map.
In the case of CNN backbones, all measures are improved when utilizing a second feature map instead of just one, i.e., excluding only the third (earliest) layer in Figs. \ref{fig:vgg}, \ref{fig:resnet}, \ref{fig:vit}. When shifting from using feature maps from two to three layers, the results are somewhat mixed; these mixed results could be attributed to the extra noise of feature maps taken earlier in the backbone's pipeline.
However, considering these results across all backbones supports the choice of utilizing three feature maps in T-TAME.
\begin{figure*}[!ht]
\centering
\includegraphics[width=\textwidth]{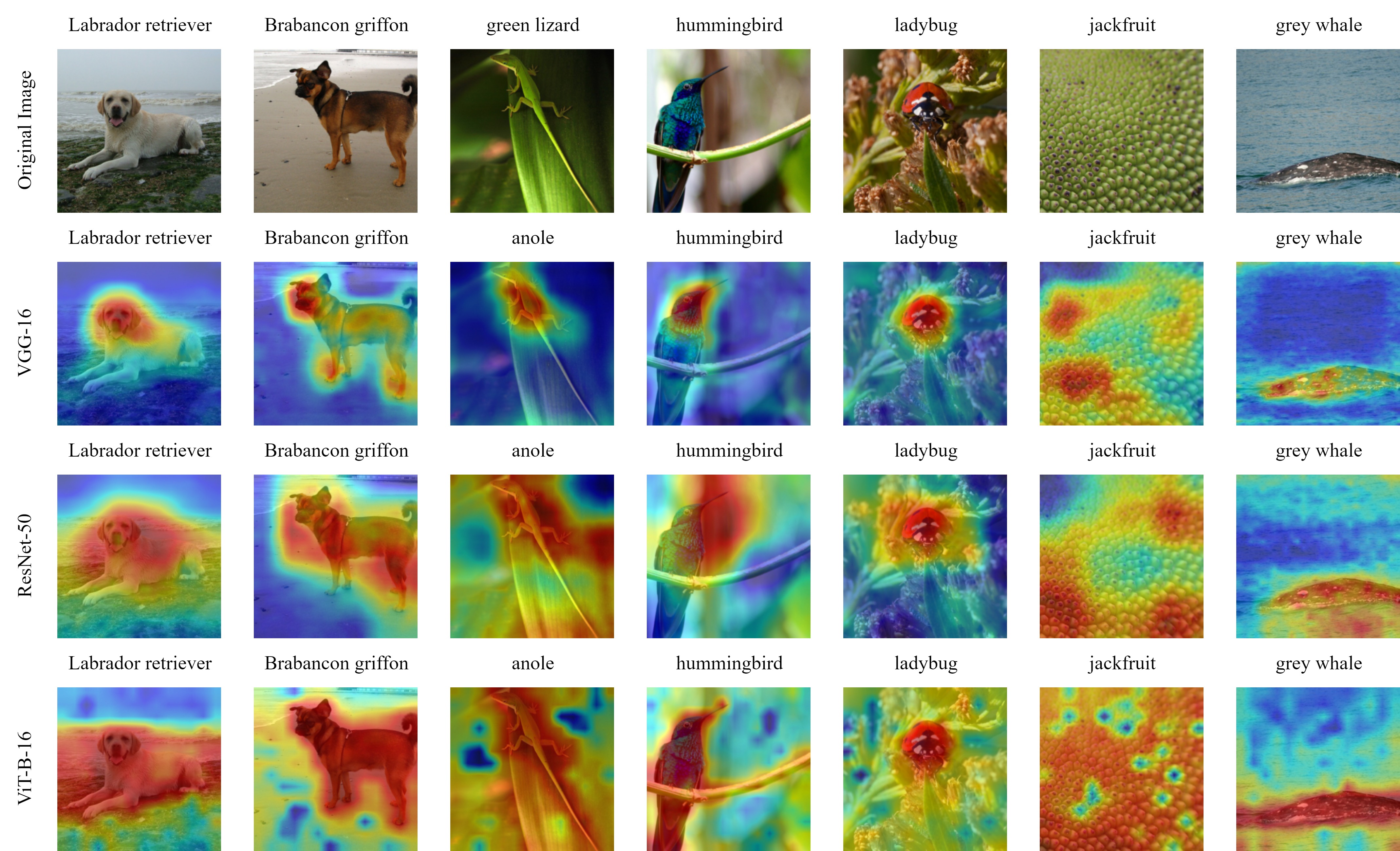}
\caption{T-TAME applied to VGG-16, ResNet-50 and ViT-B-16 backbones. We report the ground truth classes for each input image (top) and the predicted classes for each backbone (above the corresponding explanation map).
A general observation is that the explanation maps produced using the ViT-B-16 backbone attribute significance to larger image regions in comparison to the CNN backbones, highlighting the global view of the input thanks to the Transformer's Multi-head Attention layer.
}
\label{fig:backbones}
\end{figure*}
\begin{figure*}[!th]
\centering
\includegraphics[width=0.97\textwidth]{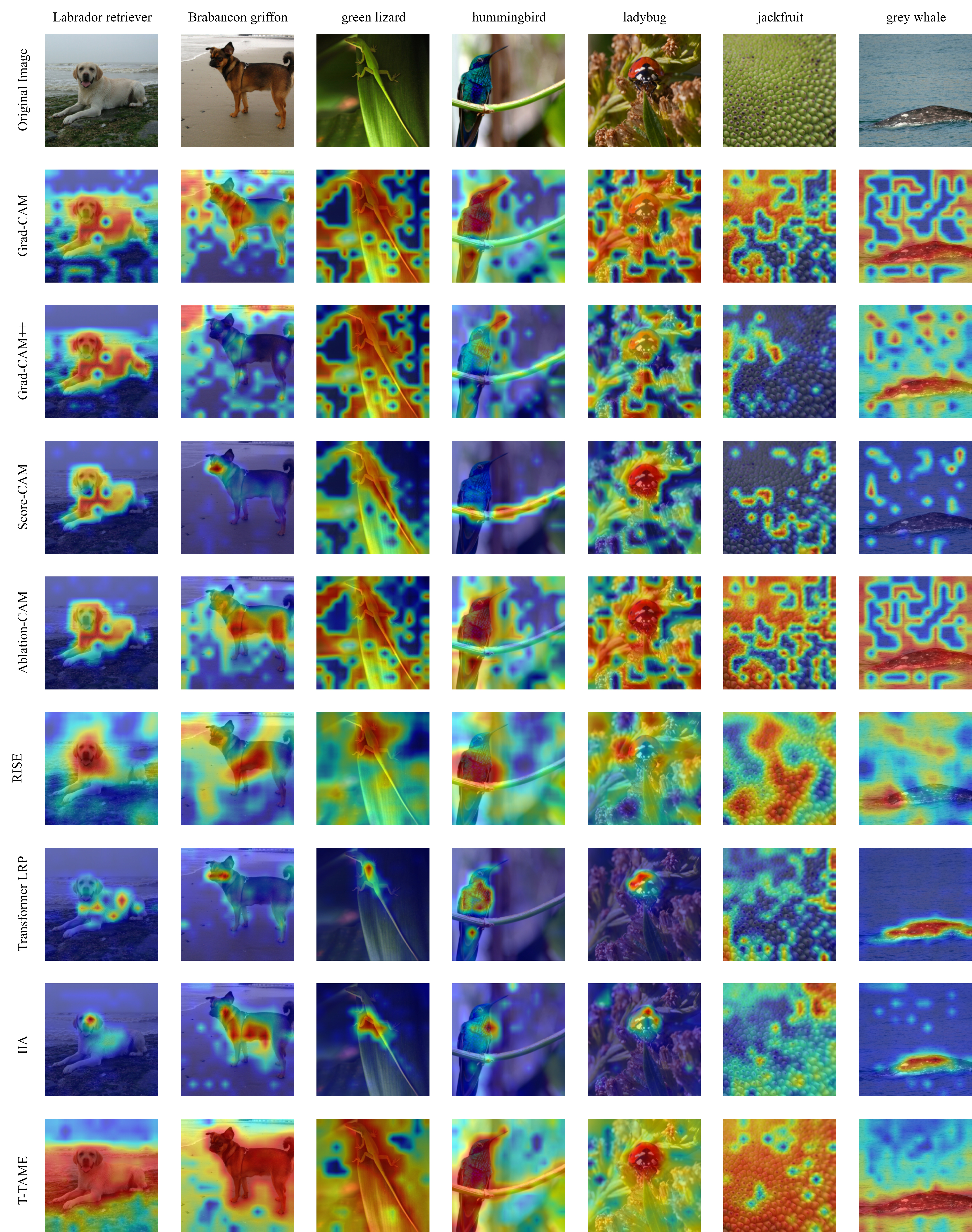}
\caption{Qualitative comparison between T-TAME and the other explainability methods of Table~\ref{tab:adic-vit} for the ViT-B-16 backbone.
We observe that T-TAME produces more activated explanation maps, demonstrating the global context used by the ViT-B-16 architecture.
}
\label{fig:comp}
\end{figure*}

\textit{Overall remarks on the attention mechanism}: We note that by omitting both the skip connection and the batch normalization in the feature branch architecture, we obtain generally better results in the case of the VGG-16 model, but this is not the case for the same architecture applied to the ResNet-50 model.
In addition, all the examined architecture variations struggle under the more challenging $v=15\%$ measures, being in most cases outperformed by the proposed T-TAME architecture;
the latter is shown to generalize the best across different backbone models.

\begin{figure*}[!ht]
\centering
\begin{tabular}{c}
     
     \includegraphics[width=\textwidth]{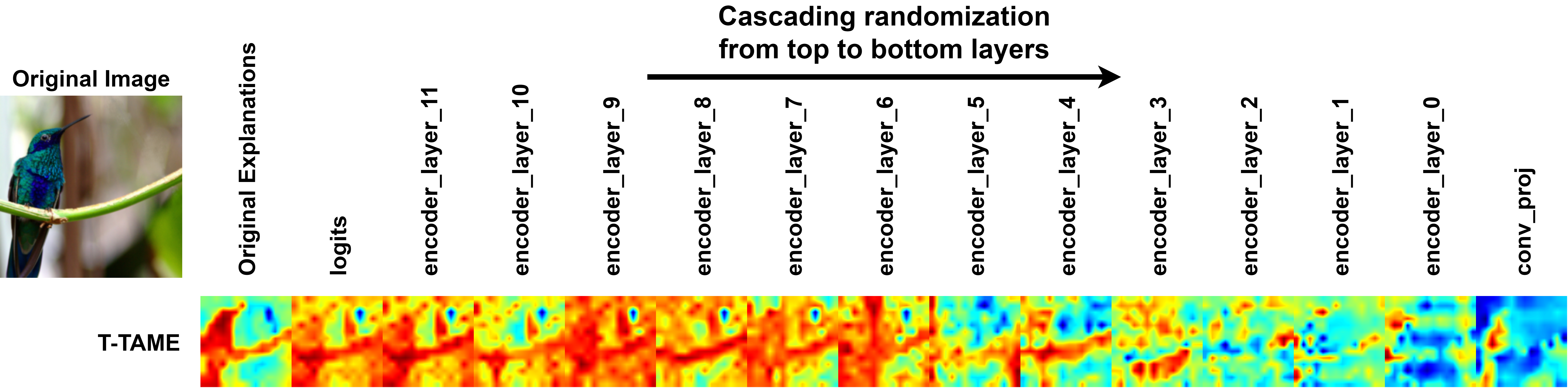} \\
     (a) \\
     \includegraphics[width=0.4\textwidth]{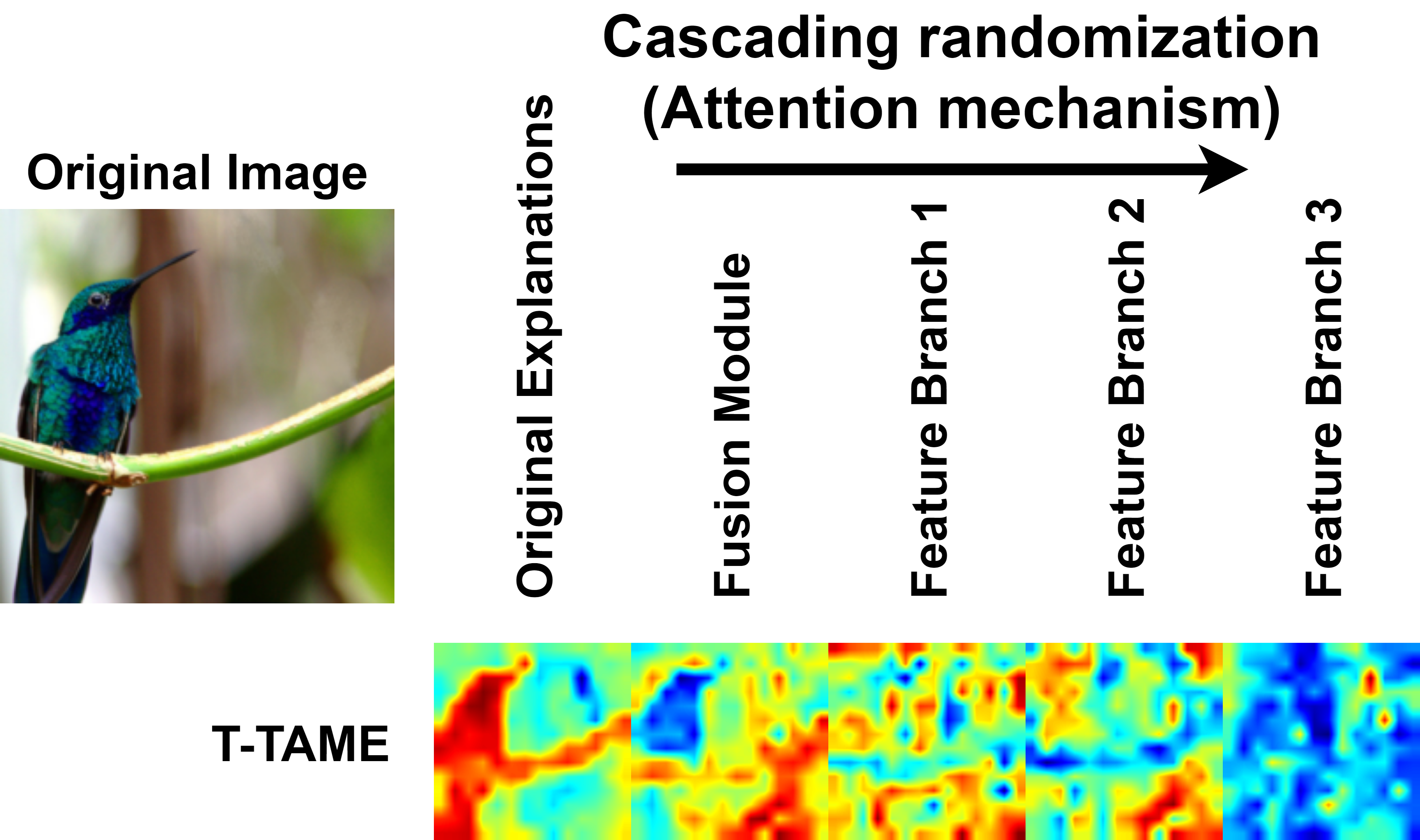} \\ 
     (b) \\
\end{tabular}

\caption{Qualitative sanity check of the proposed T-TAME method. In (a) we randomize the weights of the backbone network (ViT-B-16) in a cascading manner. In (b) we gradually randomize the attention mechanism of T-TAME. We can observe a drastic drop in the quality of the produced explanation when randomizing the backbone, starting with the logit-producing layer and finishing with the initial patch-processing convolutional layer. When randomizing the attention mechanism, the result is also a dramatic change in the produced explanation map.}
\label{fig:sanity}
\end{figure*}

\begin{figure*}[ht!]
\centering
\includegraphics[width=\textwidth]{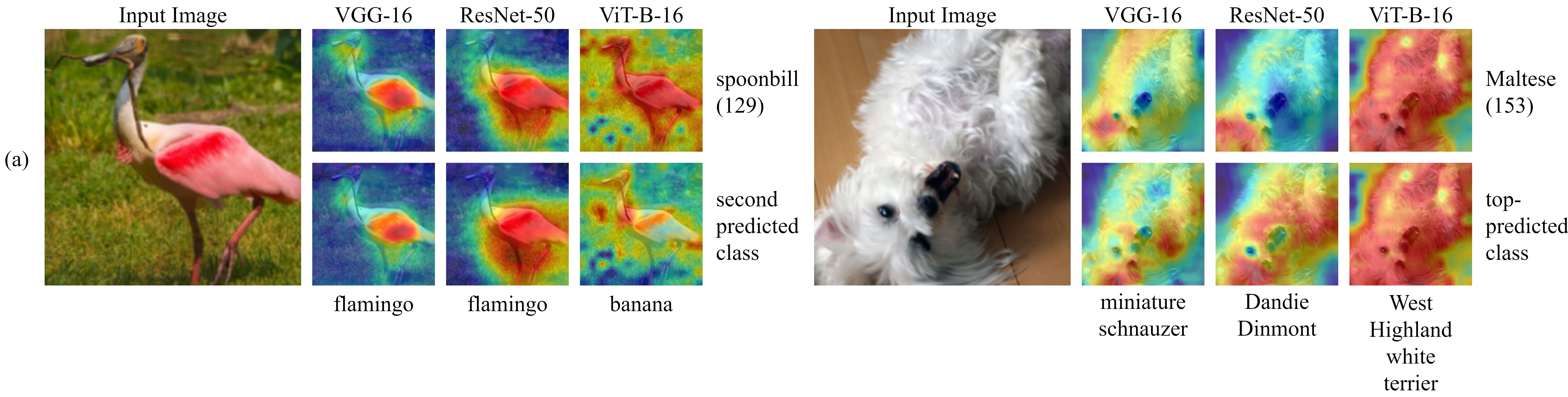} 
\caption{Counterfactual explanations for two input images.
In each case, we display six class-specific explanations for VGG-16, ResNet-50, and ViT-B-16.
The first row of explanations for each image corresponds to the image's ground-truth class, whereas the second row to the other classes: for the image on the left that is correctly classified by all three backbones, these are the second-best predictions of each backbone, while for the image on the right that is misclassified by all three backbones, these are the erroneously-predicted class of each backbone.}
\label{fig:casesa}
\end{figure*}

\begin{figure*}[ht!]
\centering
\includegraphics[width=\textwidth]{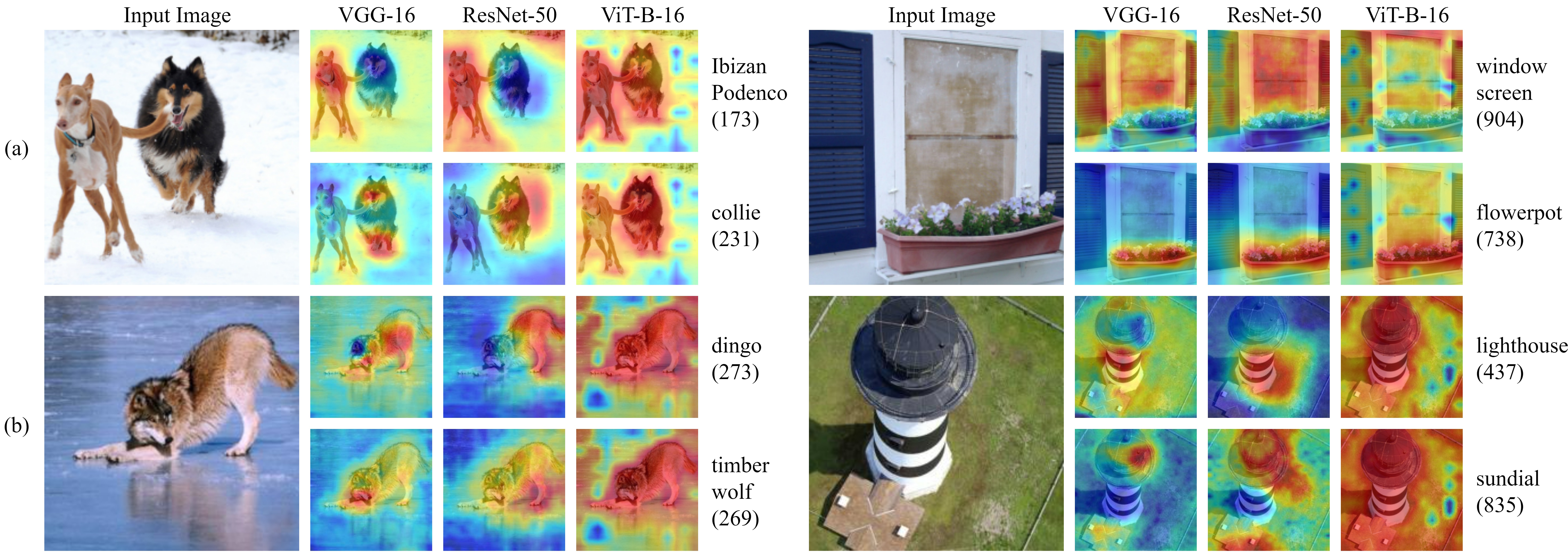} 
\caption{Explanations for four input images.
    In each case, we display six class-specific explanations, i.e., of the true (ground truth) (top) and an erroneous (bottom) class prediction of the input image, for VGG-16, ResNet-50, and ViT-B-16.
    In (a), example images with multiple classes, along with generated explanations for each respective class are depicted.
    In (b), two cases of misclassification are provided: dataset misclassification (left-side example) and model misclassification (right-side example).
    }
\label{fig:casesbc}
\end{figure*}

\subsubsection{Mismatched backbone-masking procedure combination}

As discussed in Section~\ref{sssec:norm}, CNN backbones are generally sensitive to out-of-distribution samples. Thus, in T-TAME, we introduced different procedures for masking the input with the explanation maps when working with Convolutional (Eq.~\eqref{e:cnnmasking}) or Transformer-like (Eq.~\eqref{e:vitmasking}) backbones. In this ablation experiment, we assess the effect of switching these procedures, i.e., conversely applying our CNN-specific masking procedure on the ViT backbone and our ViT-specific masking procedure on the CNN backbones. We can see in Table~\ref{tab:norm} that for the ViT-B-16 backbone, using our ViT-specific masking procedure is beneficial, especially when looking at the challenging $v=15\%$ measures. For ResNet-50, the performance differences caused by switching the masking procedure are much greater, demonstrating the sensitivity of the skip connections and of the overall ResNet architecture to out-of-distribution inputs. Similarly, in the case of the VGG-16 backbone: the degradation of performance when masking inputs using the ViT-specific procedure is clear, although less pronounced than what it was for ResNet-50. This can be attributed to the fact that the VGG-16 architecture has no skip connections: in \cite{skip-robust} it has been shown that limiting the number of skip connections improves robustness. Summarily, this ablation experiment demonstrates the importance of handling the perturbation of inputs in the case of CNN and ViT backbones differently, in agreement with what we proposed in Section~\ref{sssec:norm}. Additionally, an interesting observation is that ViT is less sensitive to the choice of masking procedure than the two examined CNNs; this is consistent with the findings of \cite{vitrobust} on the robustness of the ViT architecture to out-of-distribution samples.

\subsection{Qualitative Analysis}

An extensive qualitative analysis is performed using images from the evaluation partition of the ILSVRC 2012 ImageNet dataset.
Specifically, we present visualization examples across different backbones for the T-TAME method (Fig.~\ref{fig:backbones}); and, focusing on the ViT backbone, for T-TAME and all other compared methods of Table \ref{tab:adic-vit} (Fig.~\ref{fig:comp}). Additionally, we conduct model randomization sanity checks (following the protocol of \cite{AdebayoNIPS2018}) on the T-TAME method (Fig.~\ref{fig:sanity}).
Finally, in Subsection~\ref{sec:insights} we provide examples where the T-TAME-generated explanations can help us to gain specific insights about the backbone model and the dataset (Figs.\ref{fig:casesa} and \ref{fig:casesbc}).

\subsubsection{Qualitative comparison of T-TAME explanations across different backbones}

The qualitative differences between explanations produced using T-TAME for the VGG-16,  ResNet-50, and ViT-B-16 backbones are examined in Fig.~\ref{fig:backbones}.
We observe that explanations produced for the VGG-16 and ResNet-50 models are generally more focused on specific regions compared to the ViT-B-16 backbone, and explanations produced for the three different backbone types primarily attend to different areas of the image.
This can be explained by the fact that T-TAME is essentially trained by perturbing the original input image. ViTs are more robust to occlusions and perturbations \cite{JainICLR22}. By leveraging disjoint and spatially separate regions, ViTs retain high accuracy even when using masked inputs (see also Section~\ref{sssec:norm}).
This result suggests that VGG-16, ResNet-50, and ViT-B-16 classify images in fundamentally different ways, focusing on different features of an input image to make their predictions. The more global way in which ViT-B-16 (and Transformers, in general) interprets input images could be one of the reasons that such Transformer-based architectures perform better in the ImageNet Large Scale Visual Recognition Challenge (ILSVRC).

\subsubsection{Explanation maps for the Vision Transformer}

In Fig.~\ref{fig:comp}, explanation maps for the ViT-B-16 backbone produced using different explanation methods are depicted.
We observe that the proposed T-TAME (last row) generates the most activated explanation maps, followed by Ablation-CAM (row six) and Score-CAM (row four).
Most other methods activate only on a small, and usually a different, part of the object in the image.
For instance, observing the explanation maps in the second column of Fig.~\ref{fig:comp} concerning the Brabancon griffon, we see that all methods besides T-TAME focus on the body, on the neck and back part of the head, or the mouth and nose.

Contrarily to the other methods, the explanation maps of T-TAME tend to highlight the overall region of the object corresponding to the model-truth label, and at the same time provide the required granularity in the activation values so that the parts of the object that explain mostly the decision of the classifier are activated at a higher degree, as shown by the very good results with the AD, IC and ROAD measures (reported in Tables \ref{tab:adic-vit} and \ref{tab:road-vit}).
This shows the effectiveness of
T-TAME in revealing the long-term relations between patches captured by the ViT-B-16 multi-head attention layer and its ability to identify the salient image regions. 
Additionally, this demonstrates the importance of evaluating the various explainability methods using the AD and IC measures at multiple $v$ thresholds (Table~\ref{tab:abl}), and particularly the significance of the $v=15\%$ over the $v=100\%$ and $v=50\%$ threshold measures in judging the quality of the generated explanations.

\subsubsection{Sanity checks of T-TAME}
Sanity checks for explanation maps \cite{AdebayoNIPS2018} aim to ensure that explainability methods produce explanations that are dependent on the specific mechanism by which the backbone network processes its inputs to reach a classification decision. By randomizing the backbone network, or the dataset image-label pairs, we expect to see drastic changes in the produced explanation maps. If these changes are not observed, the method of explanation map generation does not explain the specific backbone's decision-making mechanism. It may instead simply detect image edges, or simulate other basic image filtering methods to generate superficially-convincing explanation maps. The methods used for the comparison studies (Tables~\ref{tab:adic-cnn} and \ref{tab:adic-vit}) have been observed in \cite{AdebayoNIPS2018, Chefer_2021_CVPR, petsiuk2018rise, wang2020score, ablationcam, barkan2023visual} to pass the sanity checks, so we will focus on T-TAME. We conduct two types of sanity checks on T-TAME. 

In the first case, depicted in Fig.~\ref{fig:sanity}(a), we gradually randomize the layers of the ViT-B-16 backbone network from the output layer to the input layer. We examine the effects that layer randomization has on the explanations produced for a specific image. We witness significant and abrupt differentiation between the produced explanations and the original explanation. Specifically, after randomizing the logit-producing layer, and the fifth encoder layer, we notice a major shift in the highlighted salient regions. After having randomized the entire backbone, the produced explanation bears very little resemblance to the initial explanation. This is the expected and desired result since a randomized backbone produces random results, thus no reasonable explanations for its decisions can be produced. 

In the second case, depicted in Fig.~\ref{fig:sanity}(b), we randomize the trained attention mechanism of the T-TAME method in a cascading manner (that is, this sanity check is specific to T-TAME). After randomizing the fusion module, we observe a considerable change in the produced explanation. The produced explanation map further resembles a random heatmap, as feature branches are consecutively randomized. This is again the desirable result of this sanity check, as it demonstrates that the training step of the T-TAME attention mechanism results in weights that are necessary for producing meaningful explanation maps.

\subsubsection{Example insights on ImageNet classifiers}
\label{sec:insights}
In Fig. \ref{fig:casesa}, we provide class-specific explanation maps referring to the ground truth class but also to an erroneous class, for the three examined backbones, to examine how T-TAME can assist in model interpretability. Interpretability refers to the rationale employed by a model to generate its decisions. It is different from explainability because the focus is on the model instead of a specific classification decision. 
The first image (left side example) is correctly classified by all of the examined backbones. The explanations for the second-highest predicted class, by each backbone, are also depicted.
The second image (on the right) is incorrectly classified by all of the examined backbones. The explanations for the class predicted by each model, along with the explanations for the ground truth class, are shown.
By comparing the explanation maps for adversarial classes, we can probe at the underlying decision strategy 
and possibly gain new insights for the classifier. For example, for the first image, which depicts a ``spoonbill'', in the case of the CNN backbones, the second-highest predicted class is the class ``flamingo''. These two animals share many visual characteristics, such as body shape and color. In the case of the ViT backbone, the second-highest predicted class is ``banana'', a seemingly unrelated class to the input image. Both CNNs seem to generate their decision from generic visual characteristics such as color, shape, and background. The Transformer-based architecture seems to employ a different strategy: the image has been classified as a spoonbill with high confidence, and no other class is considered possible, so the decision ``banana'' has near zero confidence. The second image (right side example) is incorrectly classified by all of the examined backbones; the top-predicted class is different for each backbone. The ground truth class's explanation map is also depicted. The initially predicted classes are all visually similar dog breeds to the ground truth class, but even for the ViT backbone the confidence in its prediction is not high: the model recognizes that classification is unclear in this instance, instead of always outputting a single prediction of high confidence.

The examples of Fig.~\ref{fig:casesbc} (a) demonstrate the potential of the explanation maps to be used for explaining multiple different classes contained in a single image, i.e., the ``Ibizan Podenco'' and ``collie'' image, and the ``window screen'' and ``flowerpot'' image. All models can clearly distinguish between the various classes contained in the images. Interestingly, the ViT backbone highlights both dogs in the first example, varying only in the intensity of the explanation map, instead of considering the second dog a negative presence in the image, as do the CNN backbones. This corroborates with our findings that ViT models interpret the input image more globally and relationally (it may be more likely for multiple animals to exist, rather than a single animal, in an image of the ImageNet dataset).

Finally, in Fig.~\ref{fig:casesbc} (b) we provide two cases of images that have been misclassified, i.e., the predicted class is not in agreement with the ground truth label of the dataset, and we use the explanations to understand what went wrong.
The first image of Fig.~\ref{fig:casesbc} (b) belongs according to its ground truth label to the ``dingo'' class (273) but is misclassified as ``timber wolf'' by all three backbones. Visual inspection reveals that the image evidently belongs to the ``timber wolf'' class, hence this is a case of dataset mislabeling; the backbone classifiers correctly focused on meaningful parts of the image to make their decisions.
The second image depicts a lighthouse.
VGG-16 misclassified this image as a ``sundial''.
Again, using the explanations generated by T-TAME, we can understand which features led the model to produce a wrong decision.
For instance, in this case, we see that for both CNN models, the ``sundial'' explanations focus on the lighthouse roof, which might resemble a sundial, explaining the erroneous classification decision of VGG-16. ViT-B-16 correctly classifies this image. The ViT-B-16 explanation does focus more on the roof as well, but it is much less concentrated on a specific region, and in this lighthouse example also focuses on the perimeter fence of the building, again showing that this classifier utilizes information from multiple parts of the image.

\section{Conclusion}
We proposed T-TAME, a novel method for generating visual explanations for deep-learning-based image classifiers.
This is accomplished by training a hierarchical attention mechanism to make use of feature maps that are extracted from multiple layers of the backbone classifier. These feature maps are appropriately transformed according to the type of the backbone network, making T-TAME compatible with both CNN and Transformer-based classifier architectures.
Experimental results verified that T-TAME clearly outperforms gradient-based and non-trainable relevance-based explainability methods, and outperforms or is on par with perturbation-based methods while, in contrast to them, it requires only a single forward pass to generate explanations.
Possible future directions include the application of T-TAME to medical image classification problems; and, the investigation of
how we could mitigate the effects of masking with low-resolution feature maps in backbones such as ResNet-50, where the output of the backbone's last stages is inevitably of low spatial resolution.

\balance
\end{document}